\documentclass[10pt, a4paper, onecolumn]{article}

\usepackage[numbers]{natbib}

\usepackage[T1]{fontenc}
\usepackage{lmodern}

\usepackage{natbib}
\usepackage[utf8]{inputenc} 
\usepackage{booktabs}       
\usepackage{amsfonts}       
\usepackage{nicefrac}       
\usepackage{microtype}      
\usepackage{xcolor}         
\usepackage{graphicx}
\usepackage{amsmath,amssymb,amsfonts}
\usepackage{bm}

\usepackage{url}
\usepackage{hyperref}

\usepackage{algorithm}
\usepackage{algorithmicx}
\usepackage{algpseudocode}

\DeclareMathOperator*{\argmax}{arg\,max}
\DeclareMathOperator*{\argmin}{arg\,min}
\usepackage{array,multirow,graphicx}

\usepackage{geometry}
\newgeometry{vmargin={15mm}, hmargin={12mm,17mm}} 

\title{Pistol: Pupil Invisible Supportive Tool to extract Pupil, Iris, Eye Opening, Eye Movements, Pupil and Iris Gaze Vector, and 2D as well as 3D Gaze.}

\author{
	Wolfgang Fuhl, Daniel Weber, Shahram Eivazi \\
	Department of Human Computer Interaction\\
	University Tübingen\\
	Tübingen, 72076 \\
	\texttt{wolfgang.fuhl@uni-tuebingen.de} \\
	\texttt{daniel.weber@uni-tuebingen.de} \\
	\texttt{shahram.eivazi@mnf.uni-tuebingen.de} \\
}

\begin{document}

	\maketitle
	
	\begin{abstract}
		This paper describes a feature extraction and gaze estimation software, named \textit{Pistol} that can be used with Pupil Invisible projects and other eye trackers in the future. In offline mode, our software extracts multiple features from the eye including, the pupil and iris ellipse, eye aperture, pupil vector, iris vector, eye movement types from pupil and iris velocities, marker detection, marker distance, 2D gaze estimation for the pupil center, iris center, pupil vector, and iris vector using Levenberg Marquart fitting and neural networks. The gaze signal is computed in 2D for each eye and each feature separately and for both eyes in 3D also for each feature separately. We hope this software helps other researchers to extract state-of-the-art features for their research out of their recordings. 
		Link: \url{https://es-cloud.cs.uni-tuebingen.de/d/8e2ab8c3fdd444e1a135/?p=%2FPISTOL&mode=list}.
	\end{abstract}

	\section{Introduction}
	\begin{figure}
		\centering
		\includegraphics[width=\textwidth]{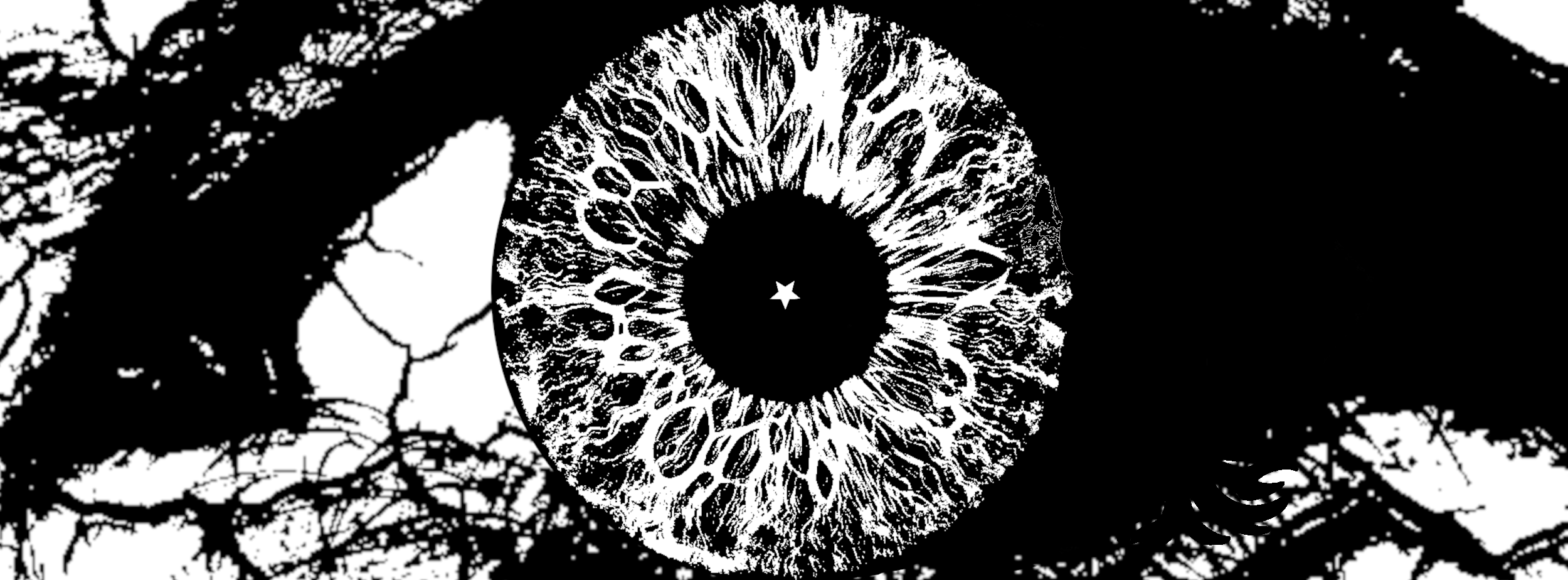}
		\caption{Part of our marker for the supportive tool Pistol.}
		\label{fig:teaser}
	\end{figure}
	Eye tracking has a variety of application areas as well as research areas in today's world. In human-machine interaction it is used as a new input signal~\cite{gardony2020eye,arslan2021eye}, in computer graphics as a constraint for the area to be rendered~\cite{walton2021beyond,meng2020eye}, in medicine using more eye features as data for self-diagnosis systems~\cite{joseph2020potential,snell2020assessment,lev2020eye} or to measure the eyes~\cite{nesaratnam2017stepping,economides2007ocular}, in the field of psychology it is used to detect neurological diseases such as Alzheimer's~\cite{davis2020eye,pavisic2021eye}, in behavioral research to evaluate expertise as well as train students~\cite{panchuk2015eye,jermann2010using}, and many more like driver monitoring in autonomous driving~\cite{liu2019gaze,shinohara2017visual} etc. This wide range of research and application areas requires that new eye features as well as more robust algorithms be easily and quickly deployable by everyone. As eye tracking itself is the focus of research and new more robust or accurate algorithms are constantly being published as well as new features, properties or signals such as pupil dilation are added, it is necessary that everyone has quick access to it to improve their products or integrate it in their research. 
	
	This poses a problem for the industry, since it must be determined long before the final product which functionalities and features will be integrated into the software~\cite{lepekhin2020adoption}. If new features are added, this is usually postponed until the next generation of eye trackers. This is due to the fact that the industry's software must be very reliable and everything should be tested multiple times, as well as automated test cases for the new parts of the software must be integrated~\cite{taylor2020operator}. Further testing must be done with other integrated components, and the software must continue to be compatible with external software components used or developed by industry partners~\cite{jiang2020design}. Another important point for the industry is the software architecture as well as also the quality of the source code. New parts must adhere to the software architecture as well as also be written in a clean and understandable way. This also delays the integration of new components, since the source code must also be checked.
	
	Research groups themselves do not have these problems, since the software architecture is usually created and extended dynamically, which however also leads to a poorer source code quality~\cite{hasselbring2020open}. Likewise, research groups have the possibility to accomplish theses for the advancement of the software, whereby no costs develop. In the case that a research group develops such software, many new algorithms are already available in working form, which makes integration much easier and faster. Also, research groups do not have contractual partners, so they do not have to maintain the output formats and integrated interfaces in the software in a prescribed format for many years.
	
	The Pistol tool presented in this paper extracts a variety of eye features such as pupil ellipse, iris ellipse, eyelids, eyeball, vision vectors, and eye-opening degree. These are needed in many applications such as fatigue or attention determination or can serve as new features in research, however these features are not provided by every eye tracker manufacturer. In addition to the extracted features, our software offers the possibility to determine the eye gaze by different methods and based on different data like the iris or the pupil with the coordinates of the center or with the vectors. This also offers new possibilities in interaction and behavioral research. We hope that this tool will be useful for scientists all over the world and also help industry to integrate new features into their software. The Pistol tool will continue to be developed in the future, and we hope to be able to integrate many scene analysis techniques.

	Our contributions to the state of the art are:
	\begin{enumerate}
		\item A free to use tool to extract a plethora of eye features for the pupil invisible offline.
		\item 2D gaze estimation per eye with multiple optimization algorithms, as well as for the pupil and the iris separately.
		\item 3D gaze estimation with both eyes using different optimization algorithms and separately for the pupil and the iris.
	\end{enumerate}
	
	\section{Related work}
	\begin{figure*}
		\centering
		\includegraphics[width=0.9\textwidth]{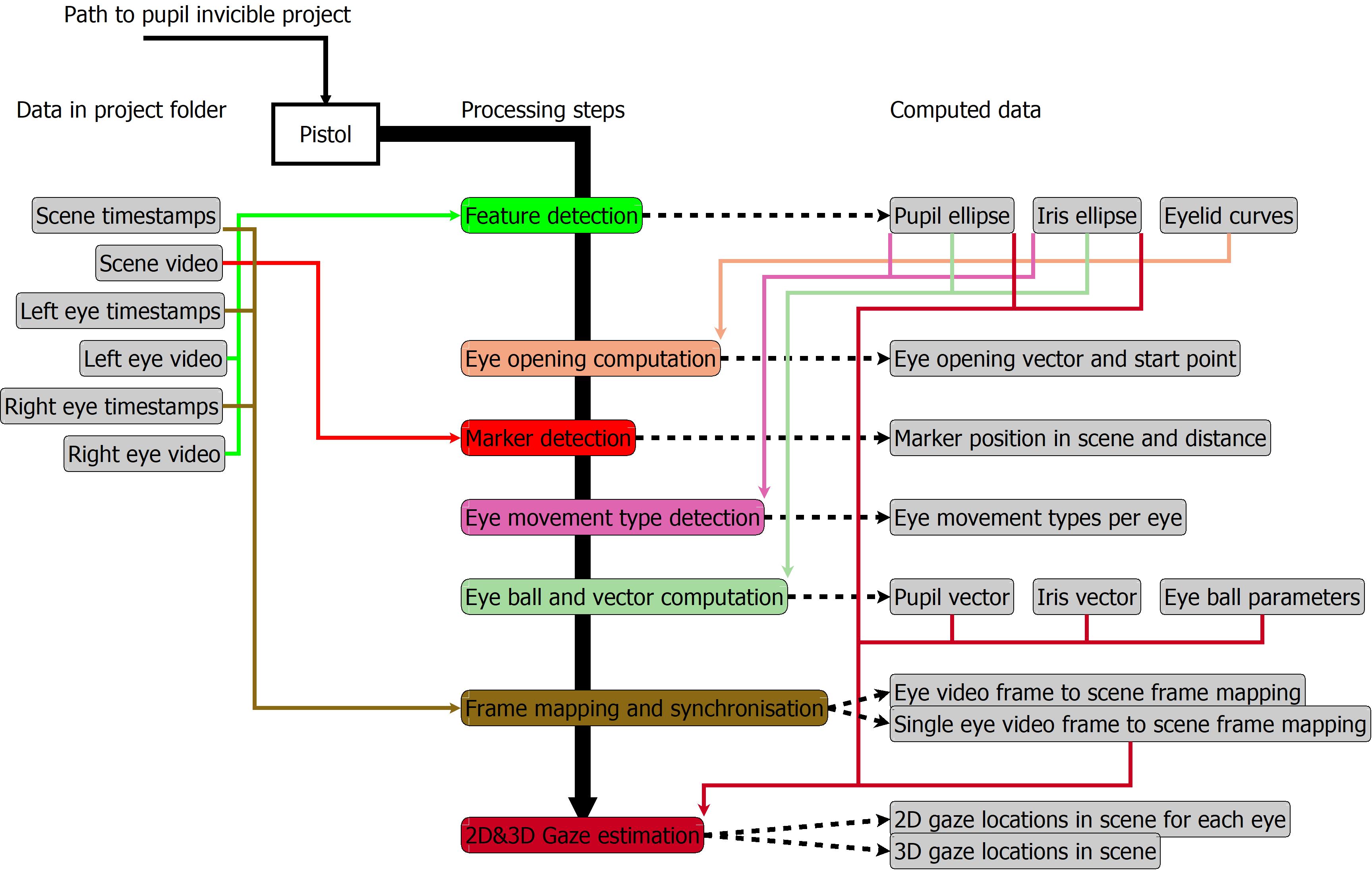}
		\caption{The workflow of Pistol. In gray are the data sources, and the single processing steps are in color. Each arrow corresponding to a data dependency is colored in the same color as the processing step.}
		\label{fig:workflow}
	\end{figure*}
	In the field of eye tracking and feature extraction, there is a wide range of related work. This is due, on the one hand, to the industry, which provides a wide range of eye trackers at different prices and with different features. On the other hand, it is due to the ever-growing research community and the growing application fields for eye tracking.
	
	From the industry there are for example the manufacturers Tobii~\cite{Tobii}, ASL~\cite{ASL}, EyeTech~\cite{EyeTech}, SR Research~\cite{SRResearch}, Eyecomtec~\cite{eyecomtec}, Ergoneers~\cite{Ergoneers}, Oculus~\cite{Oculus}, Pupil Labs~\cite{PupilLabs}, iMotions~\cite{iMotions}, VIVE~\cite{VIVE} and many more. The eye trackers differ in frame rate and in the features used for eye tracking, each having its advantages and disadvantages. For a good overview of more details, please refer to the manufacturer pages as well as survey papers that deal with this overview~\cite{rakhmatulin2020review,duchowski2002breadth,park2021technical,mao2021survey}.
	
	Science itself, has of course produced some eye trackers and systems for gaze calculation. The first one mentioned here is the EyeRec~\cite{santini2017eyerectoo} software, which can be used for online eye tracking for worn systems. It has several built-in algorithms and uses a polynomial fitting to the optical vector of the pupil. Another software dealing with pupillometry is PupilEXT~\cite{zandi2021pupilext}. This is a highly accurate extraction of the pupil shape, which can also only be performed under severe limitations and with high-resolution cameras. The software OpenEyes~\cite{li2006openeyes} offers a hardware and software solution in combination. The algorithm used for pupil detection is Starburst. For the Tobii Glasses there is also a tool~\cite{niehorster2020glassesviewer}, which allows analyzing the images offline and to apply algorithms from science to the data. For Matlab there is also a freely available interface to use Tobii Eye Tracker directly~\cite{jones2018myex}. For Tobii there also exists a custom software to improve the accuracy of the eye tracker~\cite{phan2011development}. To distribute the data of a Gazepoint Eye Tracker over a network, there is also a tool from science~\cite{hale2019eyestream}. For studies with slide shows, there is the software OGAMA~\cite{vosskuhler2008ogama}. This can be used to record mouse movements and gaze signal and then analyze them, as well as create visualizations. GazeCode~\cite{benjamins2018gazecode} is a software that allows mapping eye movements to the stimulus image. The software was developed to speed up the processing time for mapping and to improve the usability compared to commercial software like Tobii Pro Lab.
	
	The distinguishing features of our software from existing ones is that we output a variety of other features such as pupil, iris, eyelids, eye-opening, pupil vector, iris vector, eye movements and different methods for gaze estimation. Also, we determine the 3D gaze point position and would like to support a variety of eye trackers in the future if access to the image data is allowed by the license.

	\section{Method}
	Pistol is executed by calling the program with a path to the Pupil Invisible project. Then you specify the recording to be processed (psX or just X after the usage of the Pupil Player) and Pistol starts the detections and calculations. In addition, you can specify the range of marker detection to be used for calibration (based on the scene video frames). If you do not specify this range, all detected markers will be used.
	
	Figure~\ref{fig:workflow} shows the processing flow of Pistol and the data dependencies of each step. The gray boxes represent data that either exists in the project or was generated by a calculation step. Each calculation step in Figure~\ref{fig:workflow} has a unique color, with which the data dependencies are also marked. At the end of each calculation step, the data is saved to a CSV file and a debugging video is generated to check the result.
	
	In the following, we describe each processing step of our tool in detail. Some sections like the pupil, iris, and eyelid detection are combined since the fundamental algorithmic are similar.
	
	\subsection{Pupil, iris, and eyelid detection}
	\begin{figure*}
		\centering
		\includegraphics[width=0.2\textwidth]{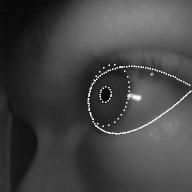}
		\includegraphics[width=0.2\textwidth]{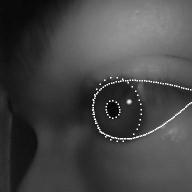}
		\includegraphics[width=0.2\textwidth]{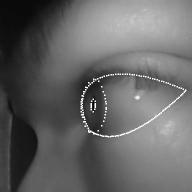}
		\includegraphics[width=0.2\textwidth]{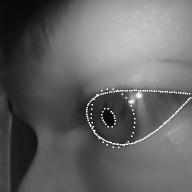}\\
		\includegraphics[width=0.2\textwidth]{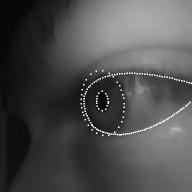}
		\includegraphics[width=0.2\textwidth]{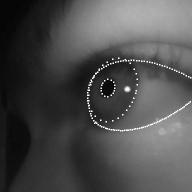}
		\includegraphics[width=0.2\textwidth]{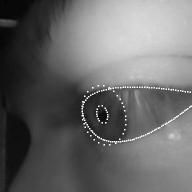}
		\includegraphics[width=0.2\textwidth]{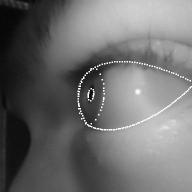}
		\caption{Exemplary detections of the pupil, iris, and eyelid for one subject. The top images are from the left eye and the bottom images are from the right eye.}
		\label{fig:pupiliriseyelid}
	\end{figure*}
	In Figure~\ref{fig:pupiliriseyelid} we show some results of pupil, iris, and eyelid detection. For detection, we use small DNNs with maximum instead of residual connections~\cite{NIPS2021MAXPROP} as well as tensor normalization and full distribution training~\cite{2021TNandFDT} to detect landmarks. Using the maximum connections allows us to use smaller DNNs with the same accuracy. The full distribution training makes our DNNs more robust, and we also need less annotated data to train them. Tensor normalization increases accuracy and is more stable than batch normalization, especially with small training datasets.  In addition, we use landmark validation as well as batch balancing from \cite{ICMV2019FuhlW} to evaluate the accuracy of the landmarks at the pixel level and discard detections if the inaccuracy is too high. To obtain the ellipses for the pupil and iris from the landmarks, we use the OpenCV~\cite{bradski2000opencv} ellipse fit. The shape of the upper and lower eyelid is approximated using cubic splines.
	
	\begin{table}[htb]
		\caption{Shows the architecture of the DNNs used for pupil, iris, and eyelid landmark detection.}
		\label{tbl:archiPIE}
		\centering
		\begin{tabular}{ll}
			\toprule
			Level & Layer \\
			\midrule
			Input & Gray scale image $192 \times 192$\\
			1 & $5 \times 5$~Convolution with depth 64\\
			2 & ReLu with tensor normalization \\
			3 & $2 \times 2$~Max pooling \\
			4 & 3 Maxium connection blocks with 64 layers and $2 \times 2$~average pooling integrated in the first block.\\
			5 & 3 Maxium connection blocks with 128 layers and $2 \times 2$~average pooling integrated in the first block.\\
			6 & 3 Maxium connection blocks with 256 layers and $2 \times 2$~average pooling integrated in the first block.\\
			7 & 3 Maxium connection blocks with 512 layers and $2 \times 2$~average pooling integrated in the first block.\\
			8 & Fully connected with 1024 neurons and ReLu\\
			Output & Fully connected with 48, 96, or 198 output neurons for pupil, iris, or eyelid landmarks in the same order.\\
			\bottomrule
		\end{tabular}
	\end{table}
	The architecture of our models is described in table~\ref{tbl:archiPIE}. With the four by three maximum interconnection blocks, our networks are between ResNet-18 and ResNet-34~\cite{he2016deep} in size. For training, we used the AdamW optimizer~\cite{loshchilov2018fixing} with parameters first momentum $0.9$, second momentum $0.999$, and weight decay $0.0001$. The initial learning rate was set to $10^{-4}$ and reduced after 1000 epochs to $10^{-5}$ in which we trained the models for an additional 1000 epochs with a batch size of 10. The training data for our models are 200 annotated images and 100 additional for validation. For data augmentation we used random noise ($0~to~0.2$ multiplied by the amount of pixels), cropping ($0.5~to~0.8$ of the resolution in each direction), zooming ($0.7~to~1.3$ of the resolution in each direction), blur ($1.0~to~1.2$), intensity shift ($-50~to~50$ added to all pixels), reflection overlay (intensity: $0.4~to~0.8$, blur: $1.0~to~1.2$, shift: $-0.2~to~0.2$), rotation ($-0.2~to~0.2$), shift ($-0.2~to~0.2$ of the resolution in each direction), and occlusion blocks (Up to 10 occlusions and with size from 2 pixels up to 50 with a fixed random value for all pixels or random values for each pixel).
	
	\subsection{Eye Opening estimation}
	\begin{figure*}
		\centering
		\includegraphics[width=0.11\textwidth]{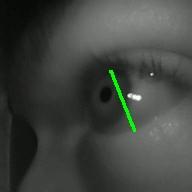}
		\includegraphics[width=0.11\textwidth]{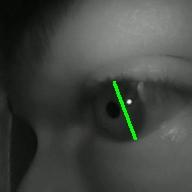}
		\includegraphics[width=0.11\textwidth]{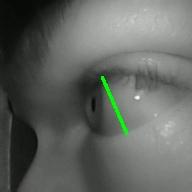}
		\includegraphics[width=0.11\textwidth]{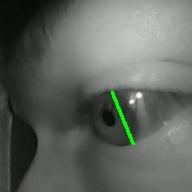}
		\includegraphics[width=0.11\textwidth]{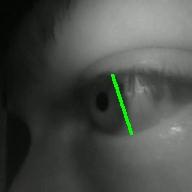}
		\includegraphics[width=0.11\textwidth]{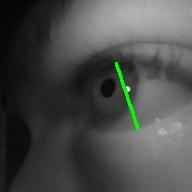}
		\includegraphics[width=0.11\textwidth]{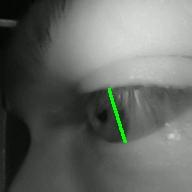}
		\includegraphics[width=0.11\textwidth]{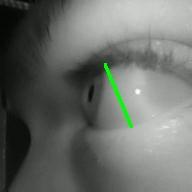}
		\caption{Exemplary images for the left (first 4) and right (last 4) eye of one subject. We compute the maximal distance along the vector between the eye corners to compensate for the off axial camera placement.}
		\label{fig:opening}
	\end{figure*}
	Figure~\ref{fig:opening} shows some results of our eye-opening degree calculation. The eye-opening is calculated using an optimization procedure and is always oriented orthogonally to the vector between the eye corners. Without the constraint of this orthogonality (simple selection of the maximum over all minimum distances of the points on the eyelid curves), the results are very erratic, because due to the perspective of the camera, parts of the eyelids are not completely visible. Also, the orthogonality to the vector between the eye corners is what is expected in a frontal view of the eye.

	

	Let $P_\text{up}$ be the set of points of the upper eyelid, $P_\text{dw}$ be the set of points of the lower eyelid, and $\overrightarrow{C}$ be the vector between the corners of the eye. 
	\begin{equation}
		\operatorname{Opening}(P_\text{up},P_\text{dw}, \overrightarrow{C}) = \argmax \argmin \lvert\overrightarrow{ud}\rvert, \quad u \in P_\text{up},~d \in P_\text{dw},~\overrightarrow{ud} \perp \overrightarrow{C},
		\label{eq:optiOpening}
	\end{equation}
Equation~\ref{eq:optiOpening} shows our optimization to compute the eyelid opening,	where $u$ and $d$ are selected elements of $P_\text{up}$ and $P_\text{dw}$ with the side condition, that the vector between the two is orthogonal to the vector between the corners of the eye, i.e. $\overrightarrow{ud} \perp \overrightarrow{C}$. For a frontal image this additional side condition is usually not necessary, but for the images from the pupil invisible the results without this side condition are unstable and not always oriented correctly regarding the eye.
	
	\subsection{Eye ball, Pupil\&Iris Vector estimation}
	\begin{figure*}
		\centering
		\includegraphics[width=0.11\textwidth]{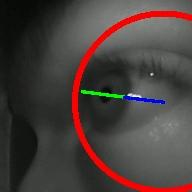}
		\includegraphics[width=0.11\textwidth]{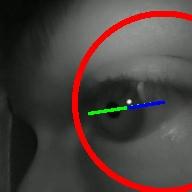}
		\includegraphics[width=0.11\textwidth]{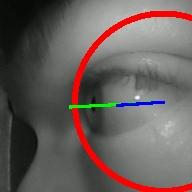}
		\includegraphics[width=0.11\textwidth]{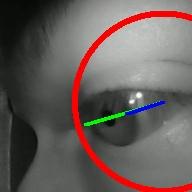}
		\includegraphics[width=0.11\textwidth]{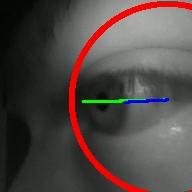}
		\includegraphics[width=0.11\textwidth]{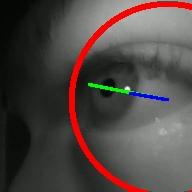}
		\includegraphics[width=0.11\textwidth]{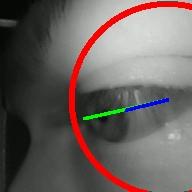}
		\includegraphics[width=0.11\textwidth]{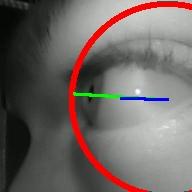}
		\caption{Exemplary images for the eyeball as well as iris and pupil vector. The eyeball is drawn in red, the iris vector in blue, and the pupil vector in green.}
		\label{fig:eyeball}
	\end{figure*}
	To calculate the eyeball and the optical vectors, we use the neural networks of \cite{NNETRA2020}. Here, several pupil ellipses are given into the neural network from which the eyeball radius and the eyeball center are calculated. As neural network, we used a network with one hidden layer consisting of 100 neurons. To calculate the optical vectors, we calculated the vector between the center and each of the pupil center and the iris center and converted it to a unit vector. 
	
	Basically the eyeball can be computed continuously in a window with this method where we sample only once over all ellipses and select the 100 most different ones. For the first update of the software an adjustable window will be integrated to continuously compute the eyeball to compensate eye tracker displacements and also the second approach from \cite{NNETRA2020} will be integrated, which allows computing an orthogonal vector from an ellipse using a neural network, which can then be used to continuously adjust the eyeball model. The choice of the method to be used is then selected via a parameter when calling it.

	\subsection{Eye movement detection}
	\begin{figure*}
		\centering
		\includegraphics[width=0.11\textwidth]{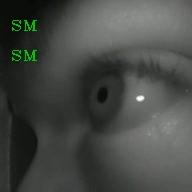}
		\includegraphics[width=0.11\textwidth]{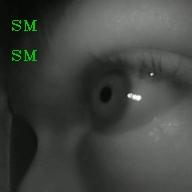}
		\includegraphics[width=0.11\textwidth]{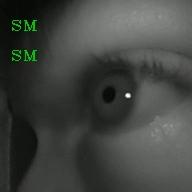}
		\includegraphics[width=0.11\textwidth]{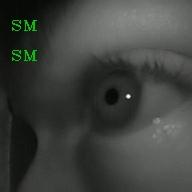}
		\includegraphics[width=0.11\textwidth]{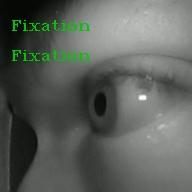}
		\includegraphics[width=0.11\textwidth]{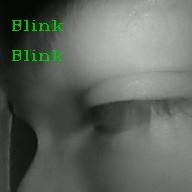}
		\includegraphics[width=0.11\textwidth]{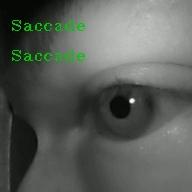}
		\includegraphics[width=0.11\textwidth]{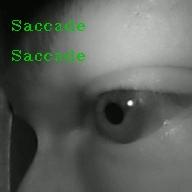}
		\caption{Exemplary detections for all eye movement types, Fixation, Saccade, Blink, and smooth pursuit. All images are from the left eye of the subject. In addition, if no feature is detected or only the eyelids are valid, but the eye is still open, the eye movement type will be marked as error. The four smooth pursuit (SM) images are from four consecutive frames with a distance of 25 frames (Eye cameras have 200FPS). The two saccade images are from two consecutive frames with a distance of 10 frames(Eye cameras have 200FPS).}
		\label{fig:eyemovement}
	\end{figure*}
	
	For the detection of eye movements, we use the angles between the pupils and iris vectors as well as the difference of the eye-opening distance. For classification, we use a neural network with a hidden layer and 100 neurons, as well as the softmax loss. The network was trained on the annotated data of five subjects. In addition, we use the validity of the extracted eye features to classify errors.
	
	Once Pistol supports multiple eye trackers, we will use the model from \cite{FCDGR2020FUHL} to segment the motion trajectories. This will be necessary because there will be significant differences in the eye trackers due to different camera placement and perspective distortion. Already with the Pupil Invisible Eye Tracker, the classification is not a linear function, as eye movements near the nose have significantly smaller distances compared to areas of the eyes that are much closer to the camera. This is due to the different depth and of course the perspective distortion of the camera lens.

	\subsection{Marker detection}
	\begin{figure*}
		\centering
		\includegraphics[width=0.2\textwidth]{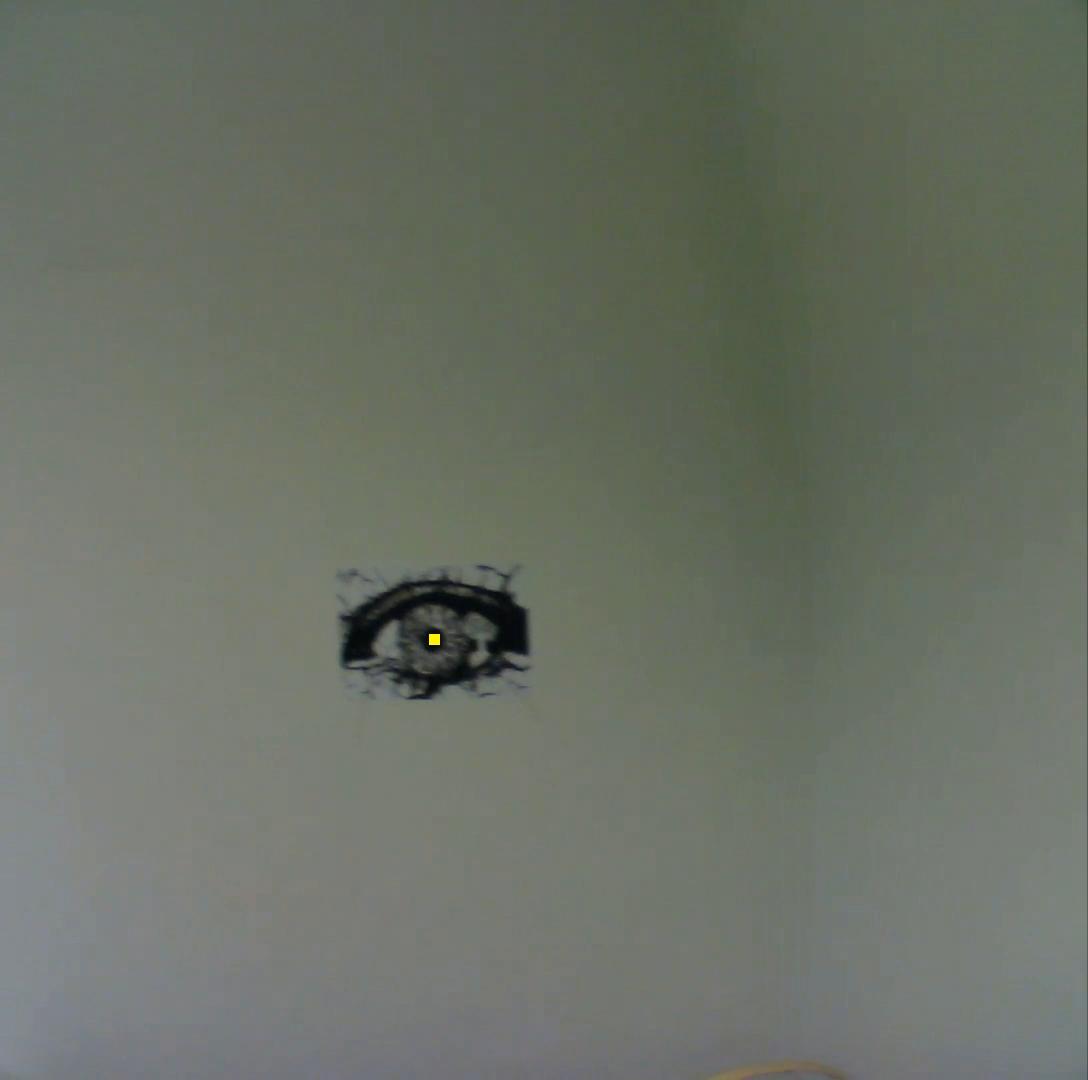}
		\includegraphics[width=0.2\textwidth]{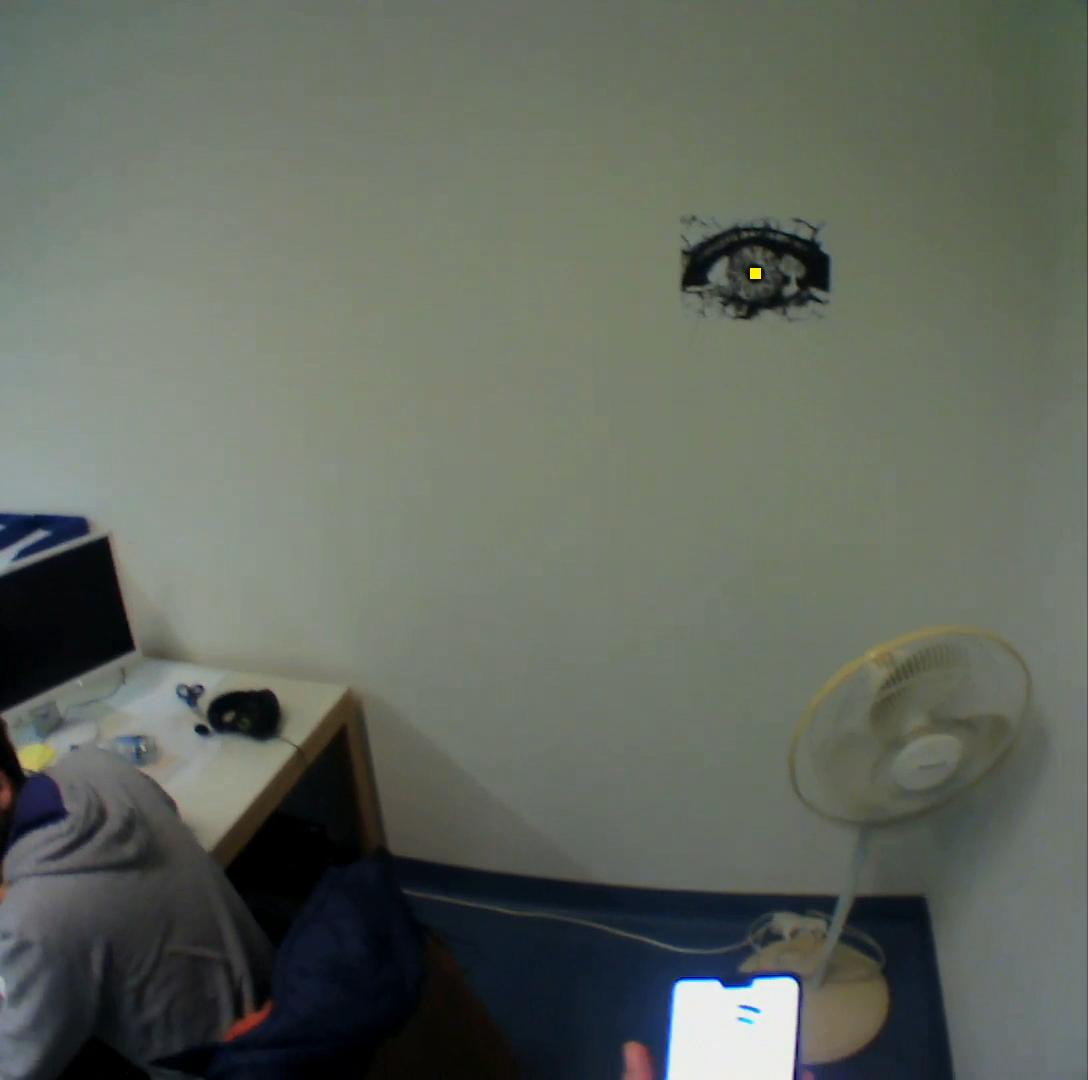}
		\includegraphics[width=0.2\textwidth]{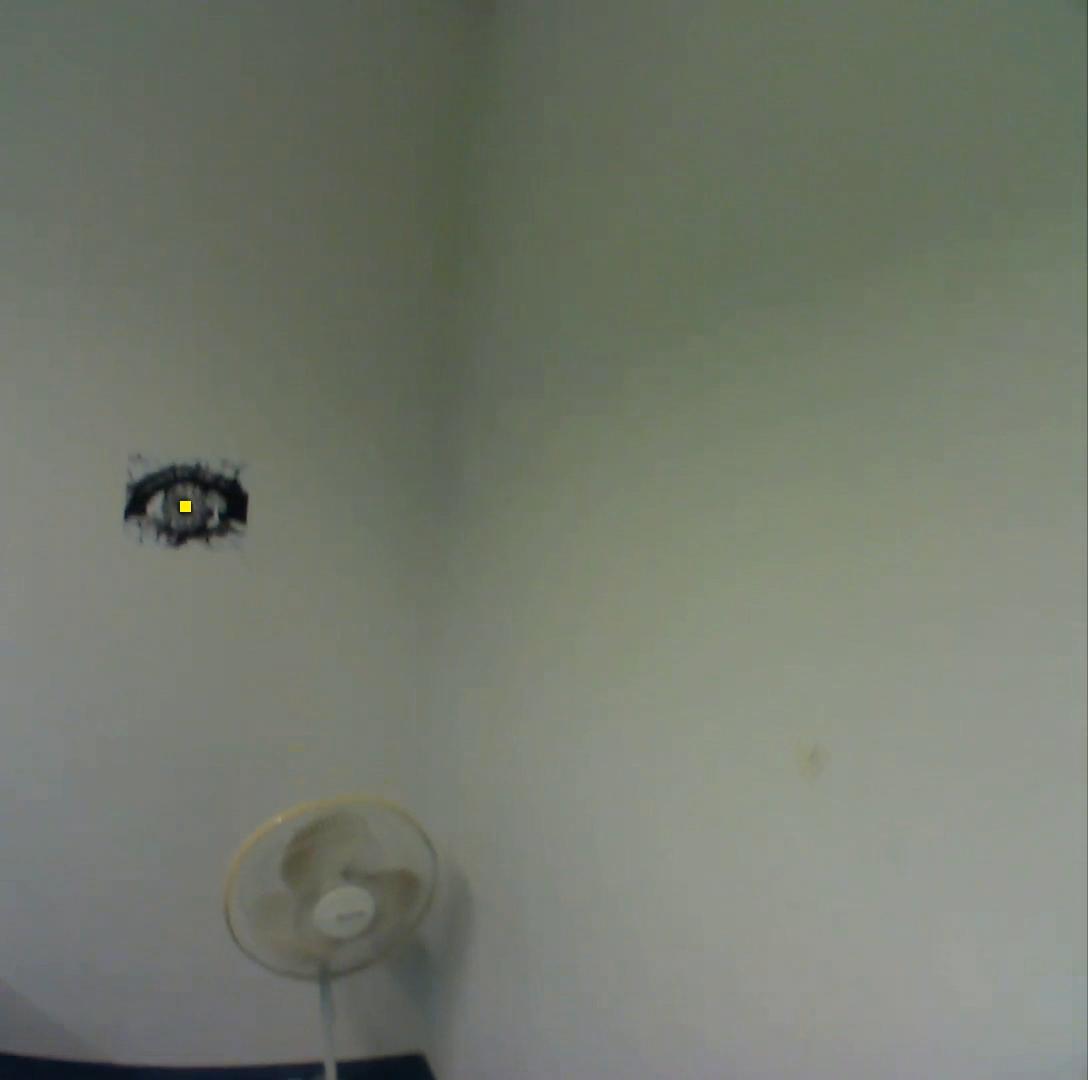}
		\includegraphics[width=0.2\textwidth]{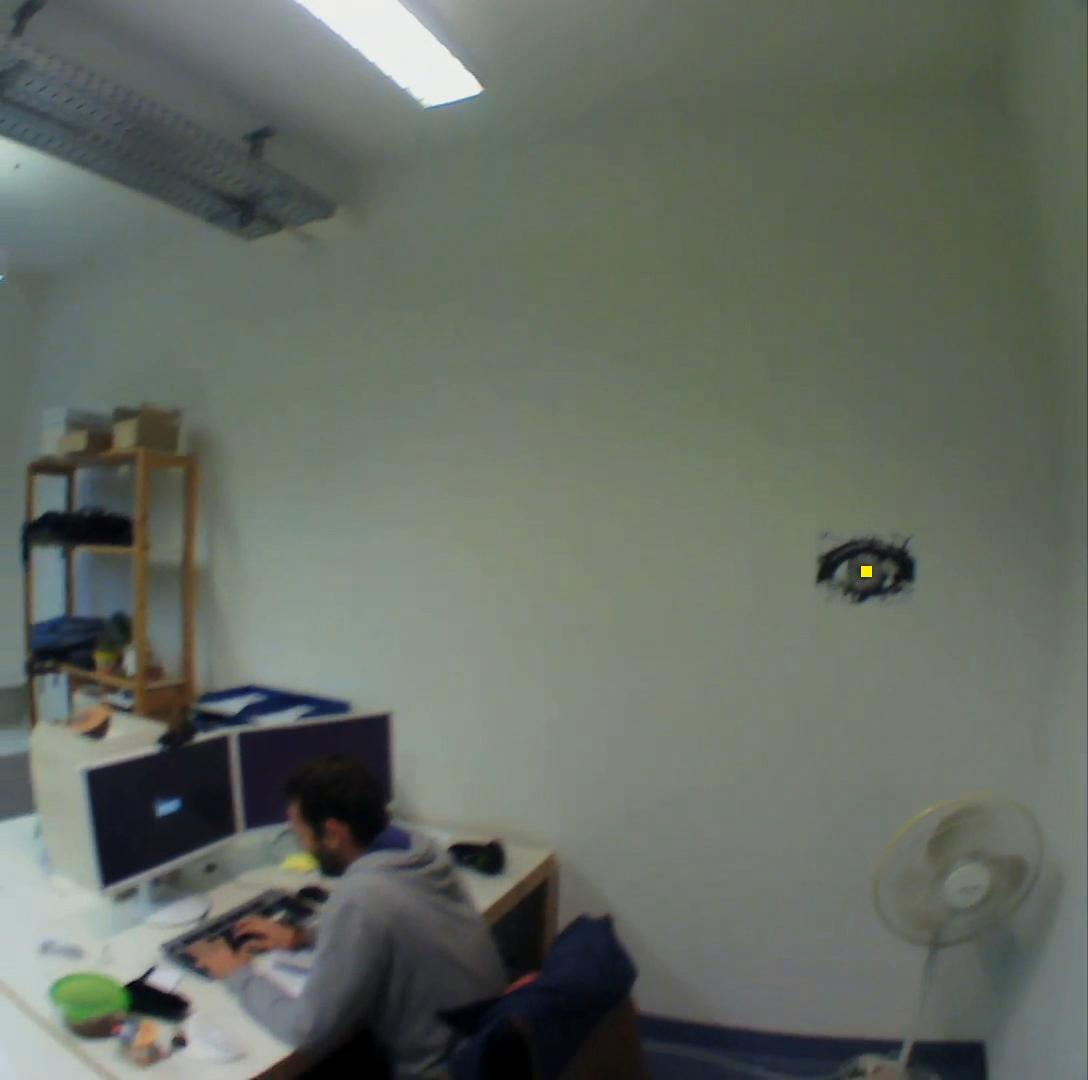}\\
		\includegraphics[width=0.2\textwidth]{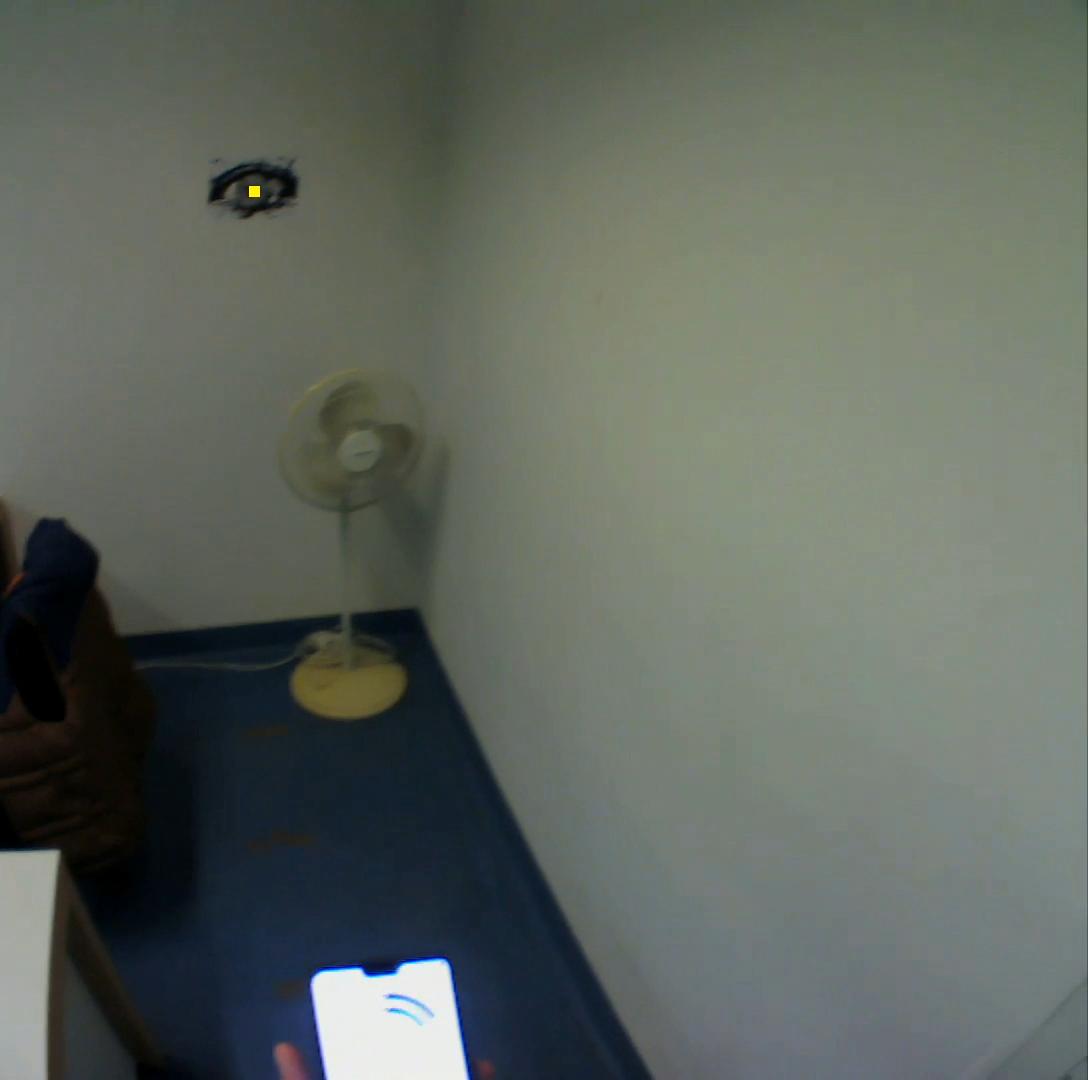}
		\includegraphics[width=0.2\textwidth]{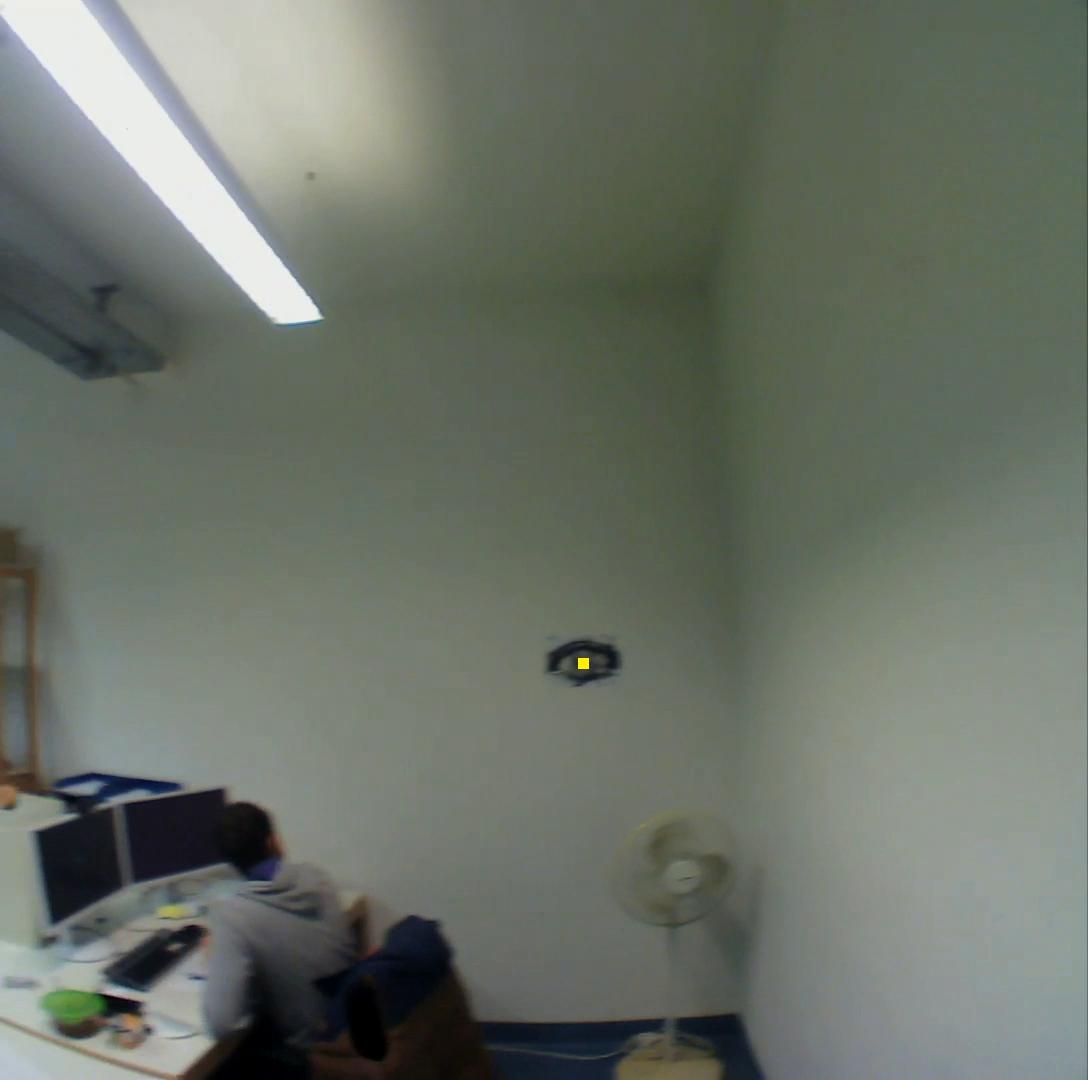}
		\includegraphics[width=0.2\textwidth]{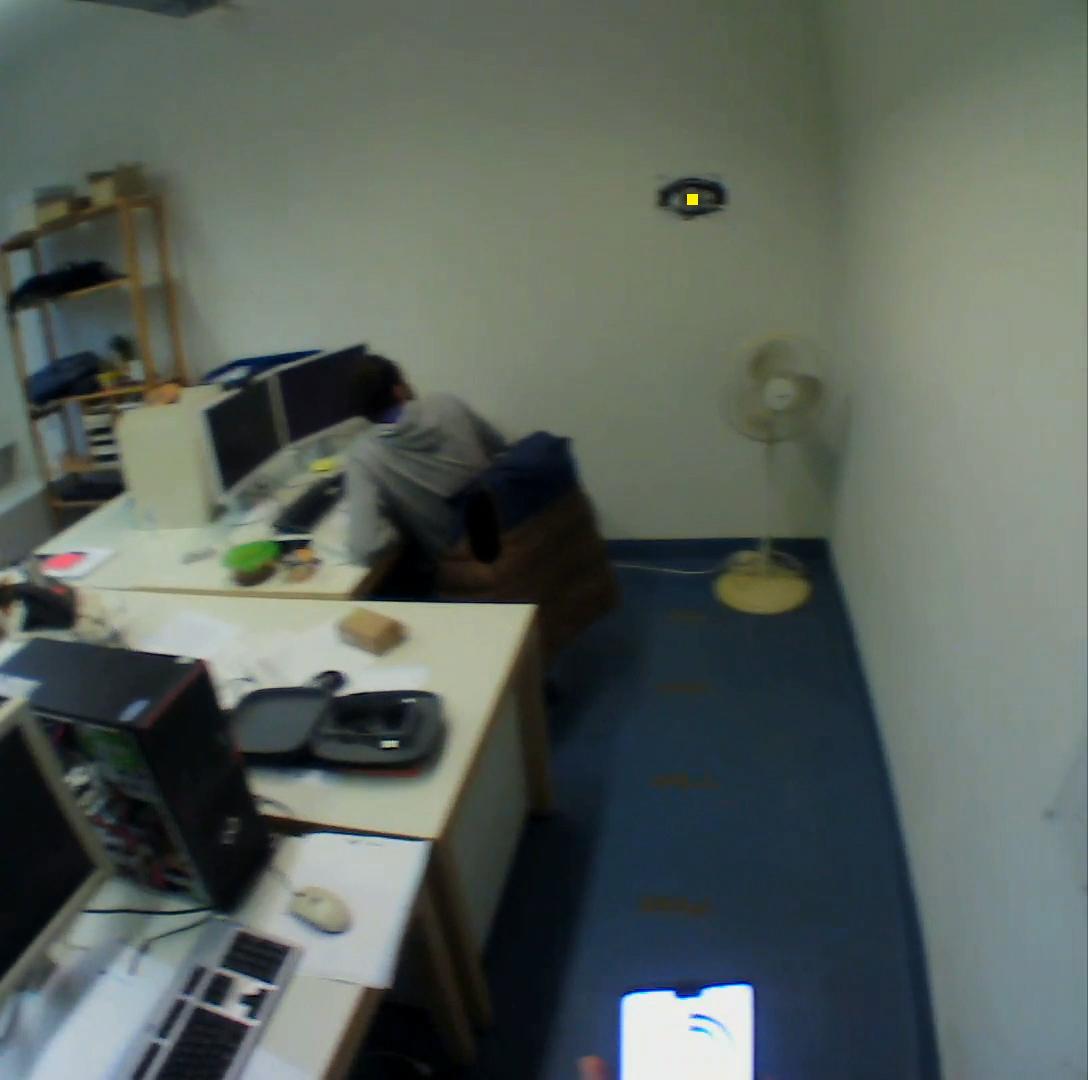}
		\includegraphics[width=0.2\textwidth]{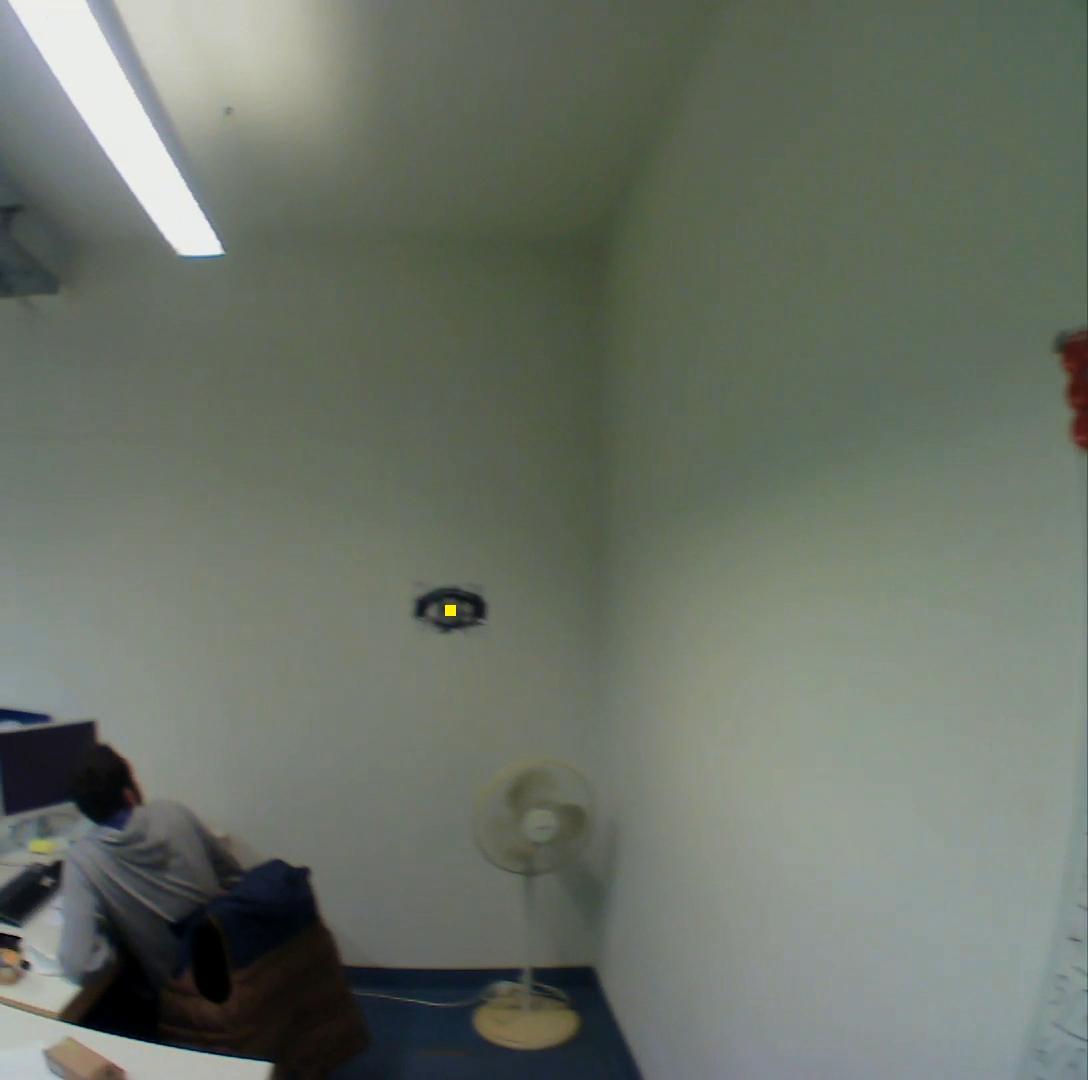}
		\caption{Exemplary images of the marker detection. The yellow dot is the estimated center while we detect the shape of the eye to estimate the area which we need to compute the distance of the marker.}
		\label{fig:marker}
	\end{figure*}
	Figure~\ref{fig:marker} shows results of our marker detection, where the center is shown as a yellow dot. Our marker detection must be able to detect the marker over different distances and should calculate the center as accurately as possible. To accomplish this in a reasonable time, we decided to use two DNNs. The first DNN gets the whole image scaled to $400 \times 400$ pixel. From this, the DNN generates a heatmap with a resolution of $50 \times 50$ pixel. The maxima in this heatmap are then used as the starting position for the second DNN, which extracts a $120 \times 120$ pixel area from the original image and performs landmark detection with validation from \cite{ICMV2019FuhlW} here. With the validation signal, we again reject marker positions which are too inaccurate.

	\begin{table}[htb]
		\caption{Shows the architectures of the DNNs used for the marker detection.}
		\label{tbl:archmarker}
		\centering
		\begin{tabular}{lll}
			\toprule
			Level & Coarse detector & Fine detector\\
			\midrule
			Input & Gray scale image $400 \times 400$ & RGB image $120 \times 120$\\
			1 & $5 \times 5$~Convolution with depth 32 & $5 \times 5$~Convolution with depth 32\\
			2 & ReLu with tensor normalization & ReLu with tensor normalization\\
			3 & $2 \times 2$~Max pooling & $2 \times 2$~Max pooling \\
			4 & $5 \times 5$~Convolution with depth 64 & 1 Maxium connection block with 64 layers and $2 \times 2$~average pooling\\
			5 & ReLu with tensor normalization & 1 Maxium connection block with 128 layers and $2 \times 2$~average pooling\\
			6 & $3 \times 3$~Convolution with depth 1 & 1 Maxium connection block with 256 layers and $2 \times 2$~average pooling\\
			7 & $4 \times 4$~Average pooling & Fully connected with 512 neurons and ReLu\\
			8 & & Fully connected with 48 output neurons for the landmarks\\
			\bottomrule
		\end{tabular}
	\end{table}
	Table~\ref{tbl:archmarker} shows the architectures of our DNNs. Both DNNs use tensor normalization~\cite{2021TNandFDT}. The maximum connections~\cite{NIPS2021MAXPROP} and full distribution training~\cite{2021TNandFDT} were also used in the fine detector. To train the DNNs we annotated 100 images (60 for training and 40 for validation) where the coarse DNN used the entire image for training and the fine DNN used only the region of the marker in a $120 \times $120 pixel area.
	
	For data augmentation we used random noise ($0~to~0.2$ multiplied by the amount of pixels), cropping ($0.5~to~0.8$ of the resolution in each direction), zooming ($0.7~to~1.3$ of the resolution in each direction), blur ($1.0~to~1.2$), intesity shift ($-50~to~50$ added to all pixels), reflection overlay (intesity: $0.4~to~0.8$, blur: $1.0~to~1.2$, shift: $-0.2~to~0.2$), rotation ($-0.2~to~0.2$), shift ($-0.2~to~0.2$ of the resolution in each direction), and occlusion blocks (Up to 10 occlusions and with size from 2 pixels up to 50\% of the image with a fixed random value for all pixels or random values for each pixel). This data augmentation was used for both DNNs.
	
	For training of the fine detector, we used the AdamW optimizer~\cite{loshchilov2018fixing} with parameters first momentum $0.9$, second momentum $0.999$, and weight decay $0.0001$. The initial learning rate was set to $10^{-3}$ and reduced after 1000 epochs to $10^{-4}$ in which we trained the models for an additional 1000 epochs with a batch size of 10.  
	
	The coarse detector was trained with SGD~\cite{rumelhart1986learning} and the parameter's momentum $0.9$ and weight decay $0.0005$. The initial learning rate was set to $10^{-1}$ and reduced after 1000 epochs to $10^{-2}$ in which we trained the models for an additional 1000 epochs with a batch size of 10.

	\subsection{2D \& 3D Gaze estimation}
	\begin{figure*}
		\centering
		\includegraphics[width=0.2\textwidth]{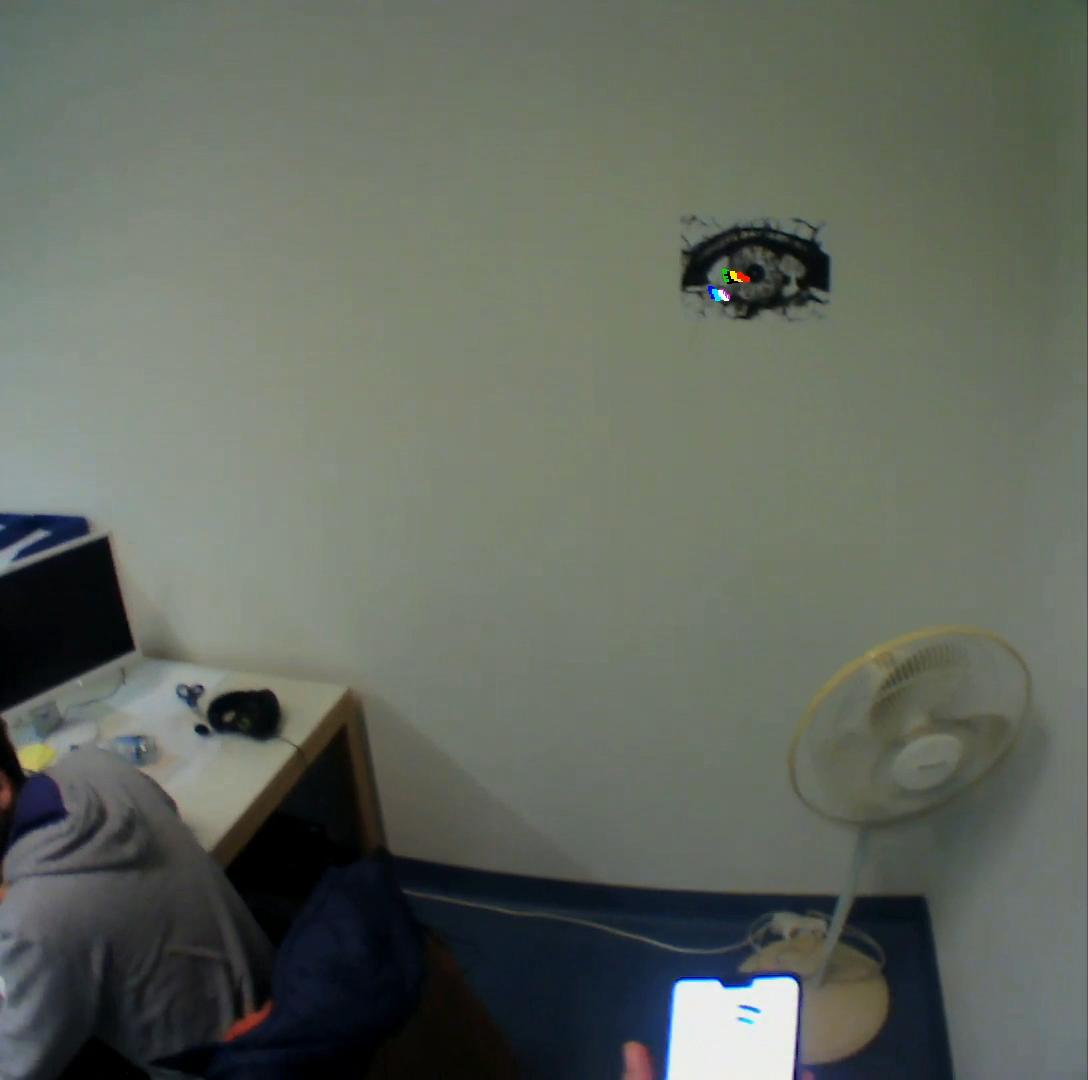}
		\includegraphics[width=0.2\textwidth]{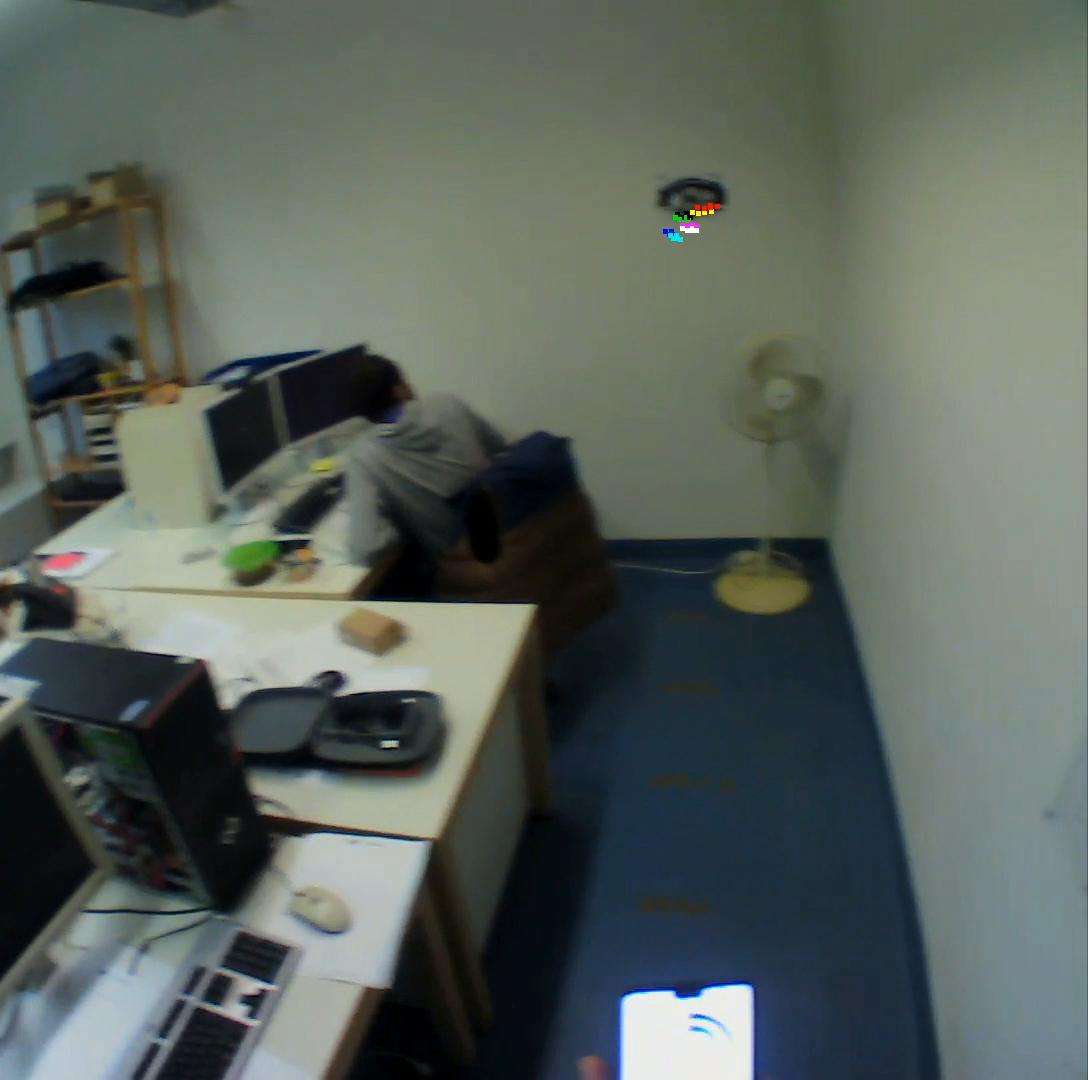}
		\includegraphics[width=0.2\textwidth]{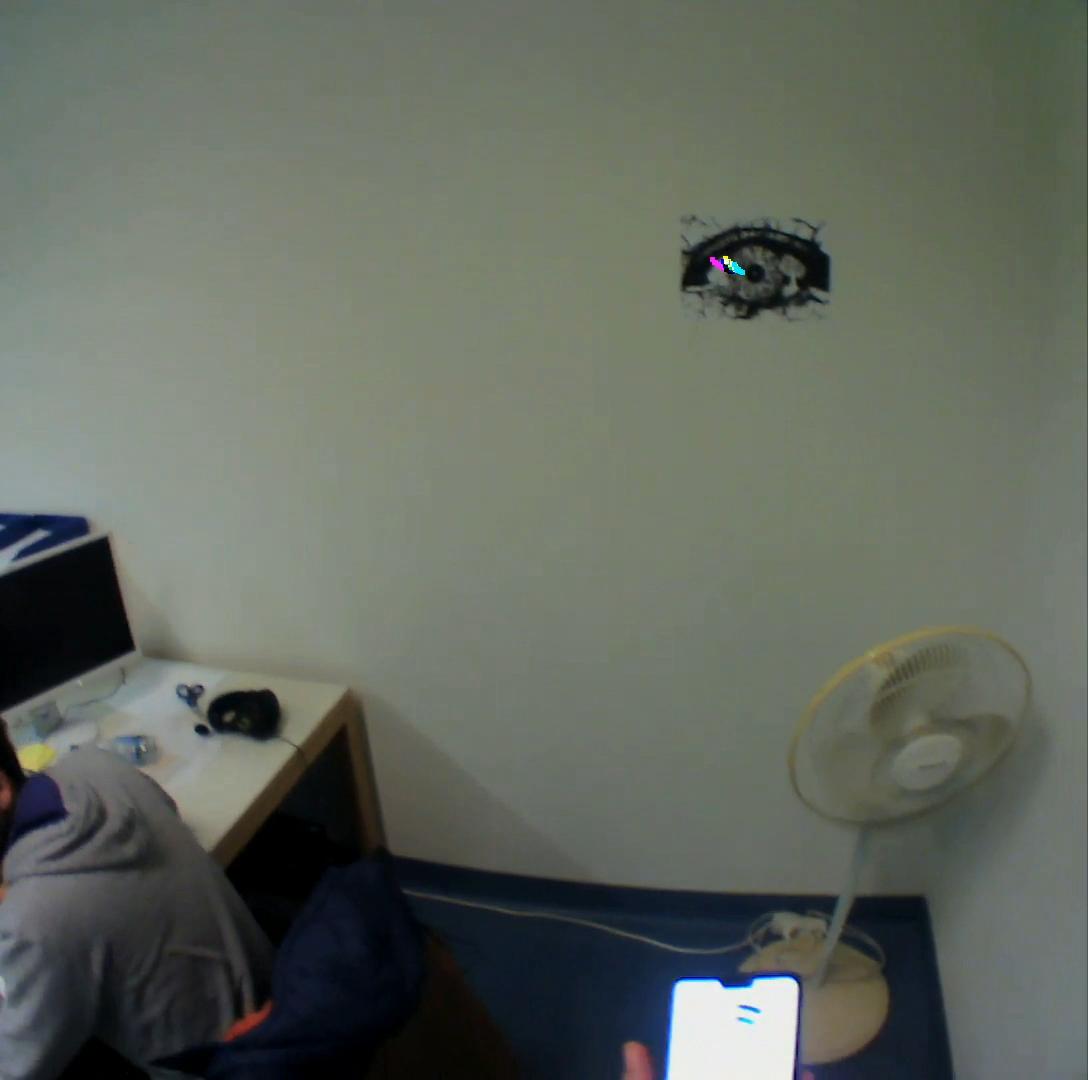}
		\includegraphics[width=0.2\textwidth]{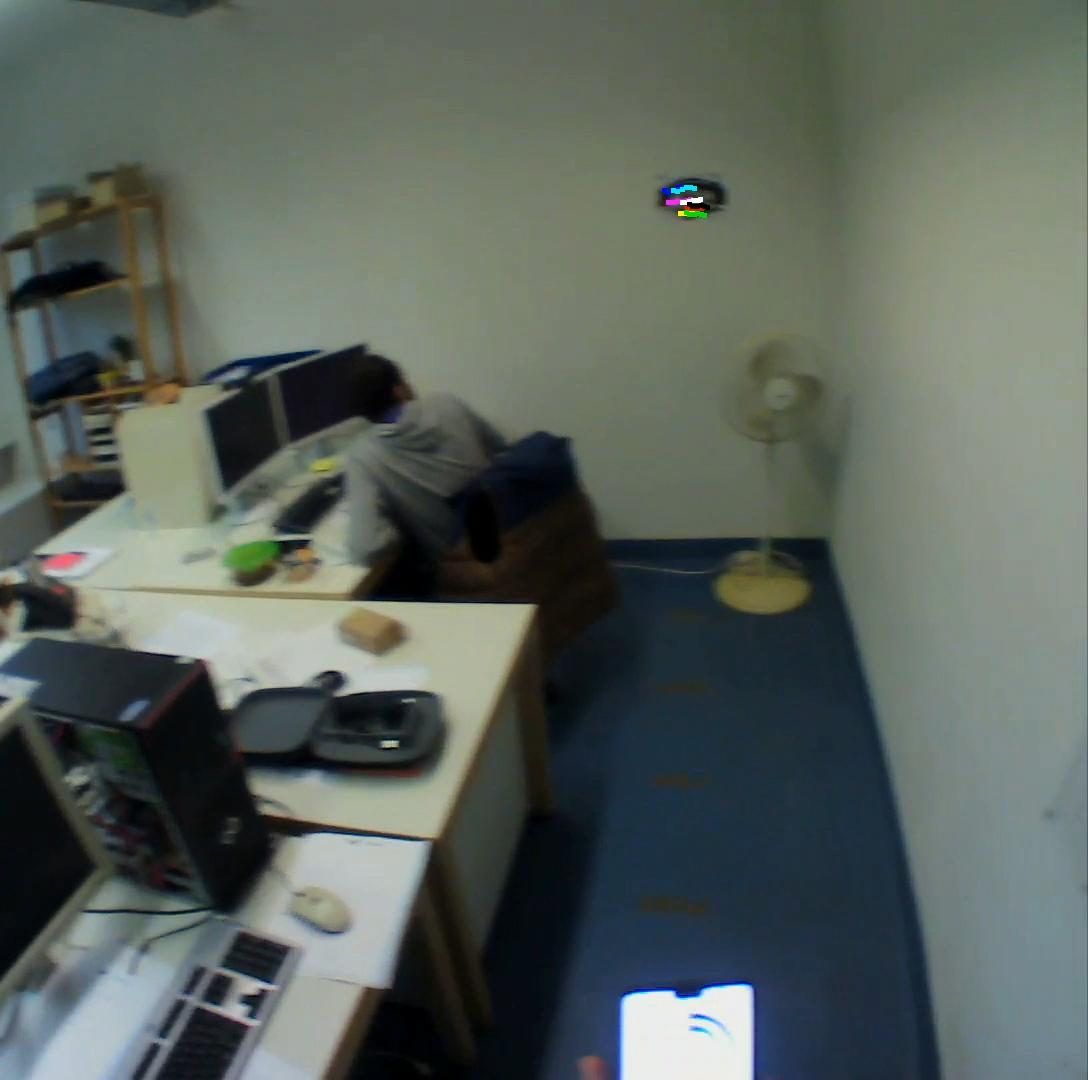}
		\includegraphics[width=0.2\textwidth]{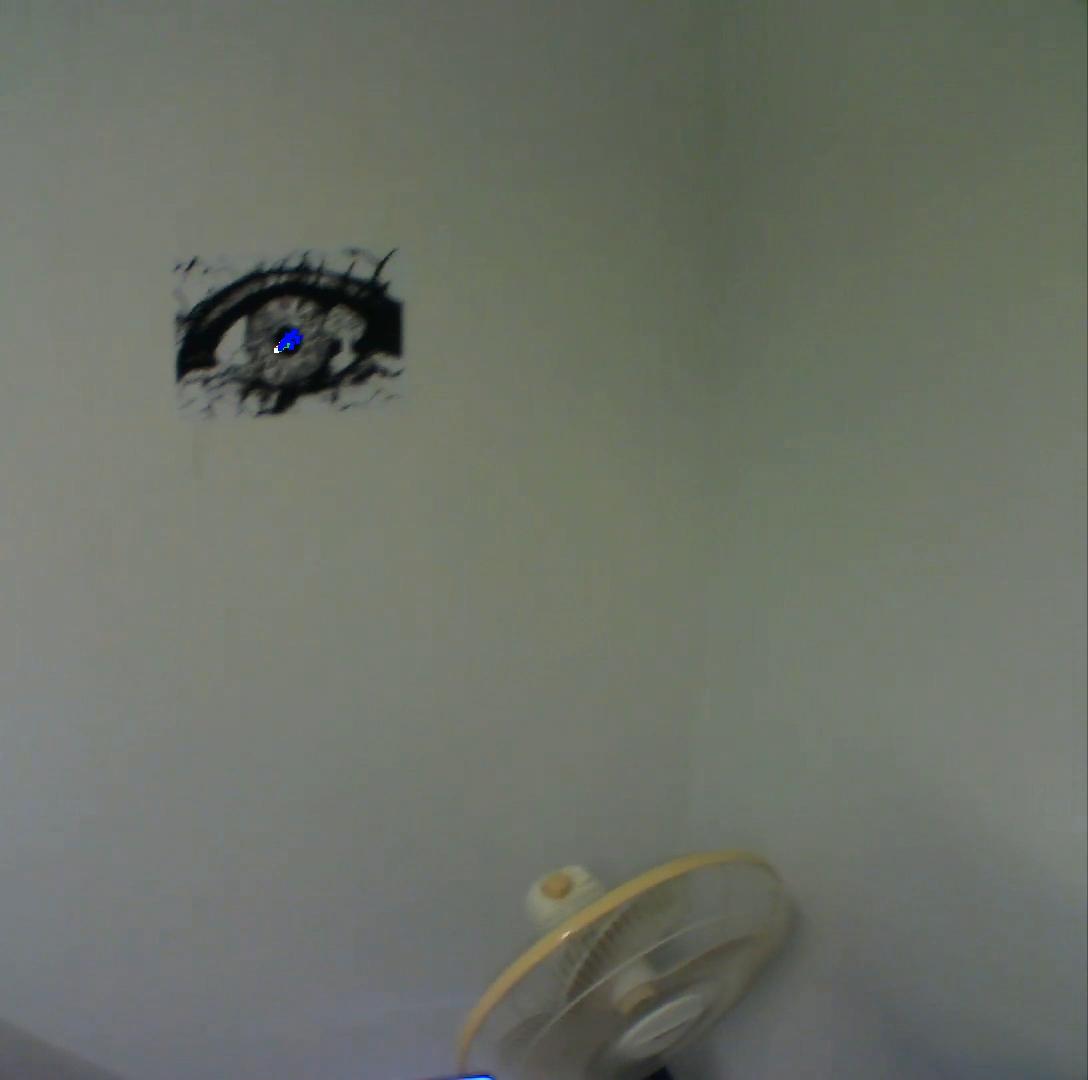}
		\includegraphics[width=0.2\textwidth]{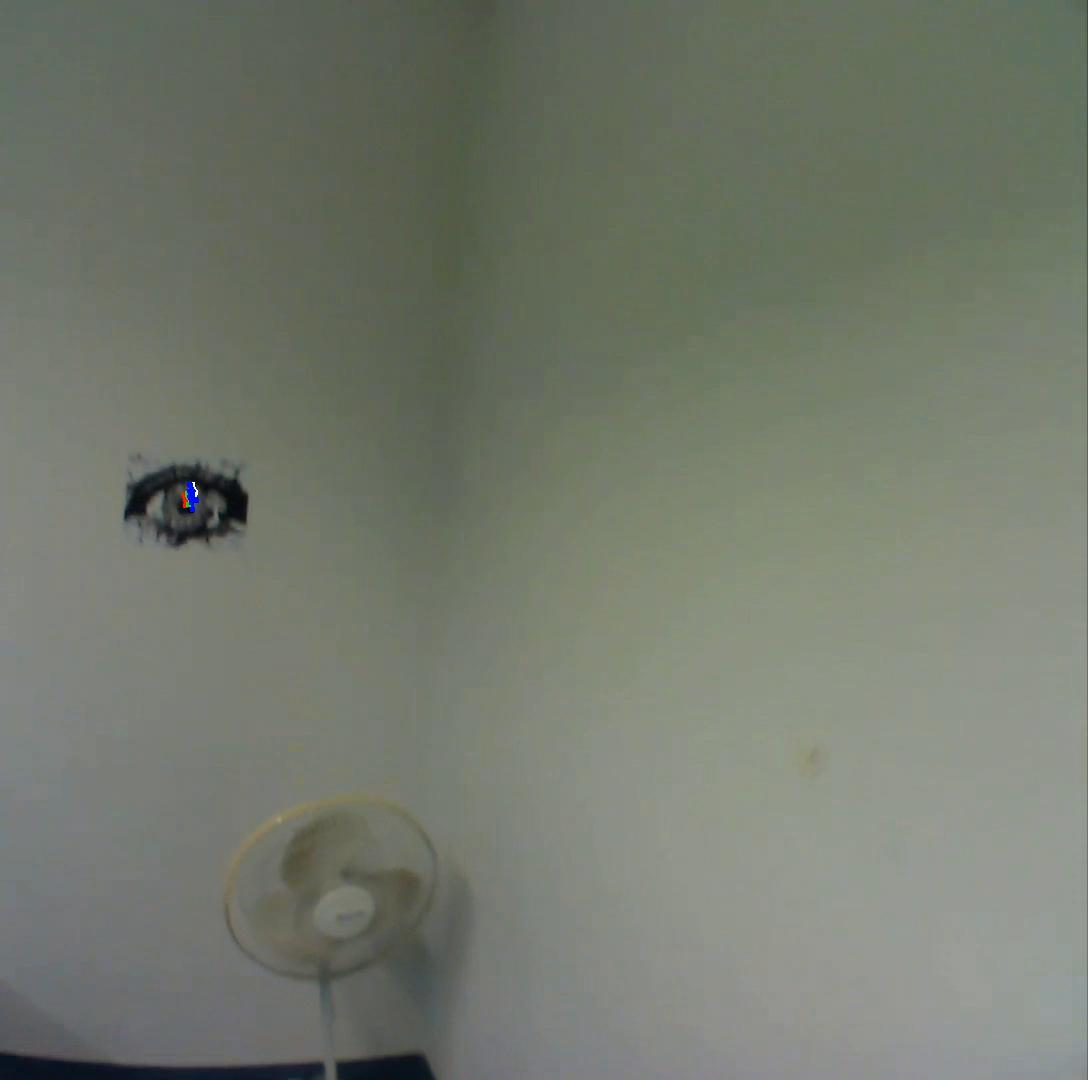}
		\includegraphics[width=0.2\textwidth]{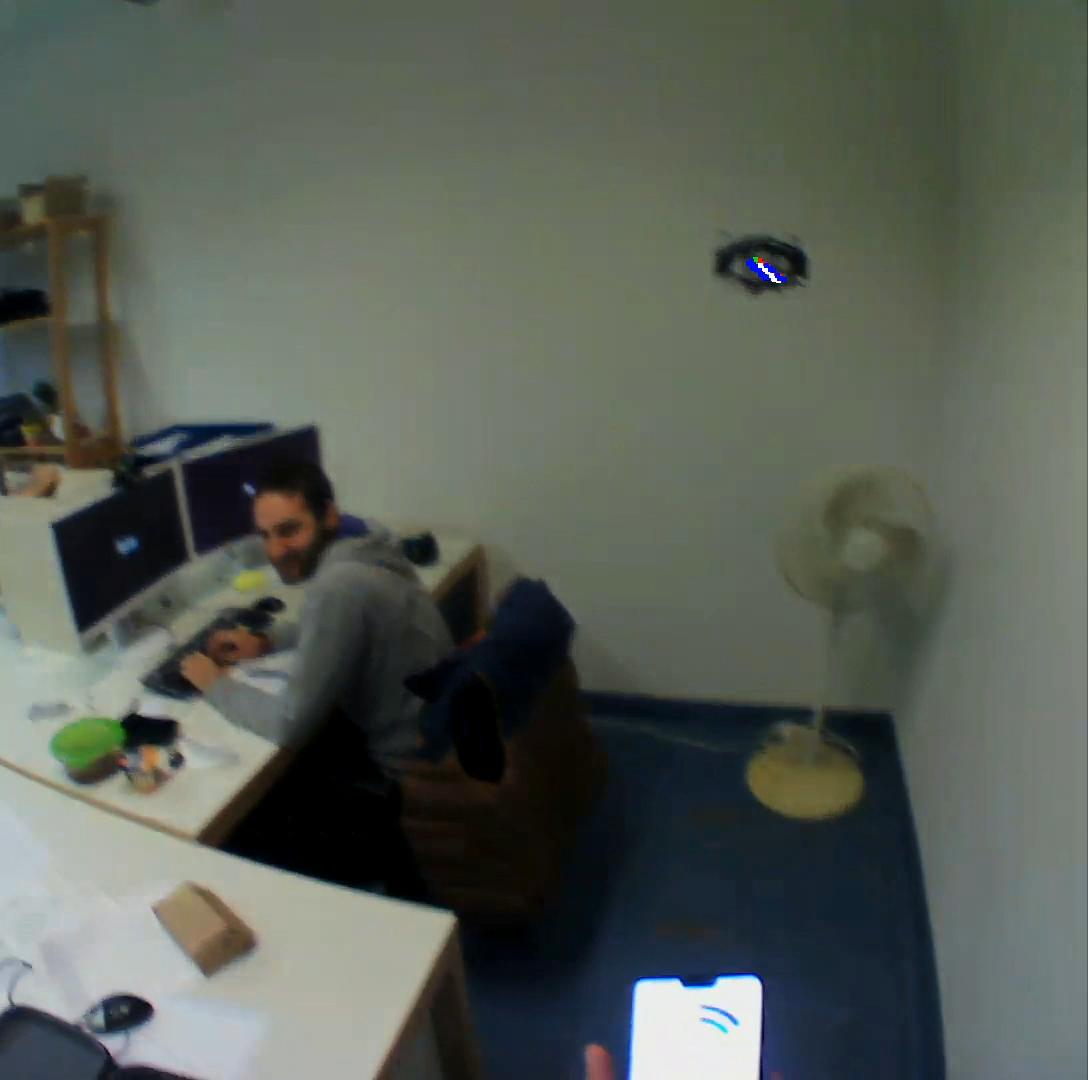}
		\includegraphics[width=0.2\textwidth]{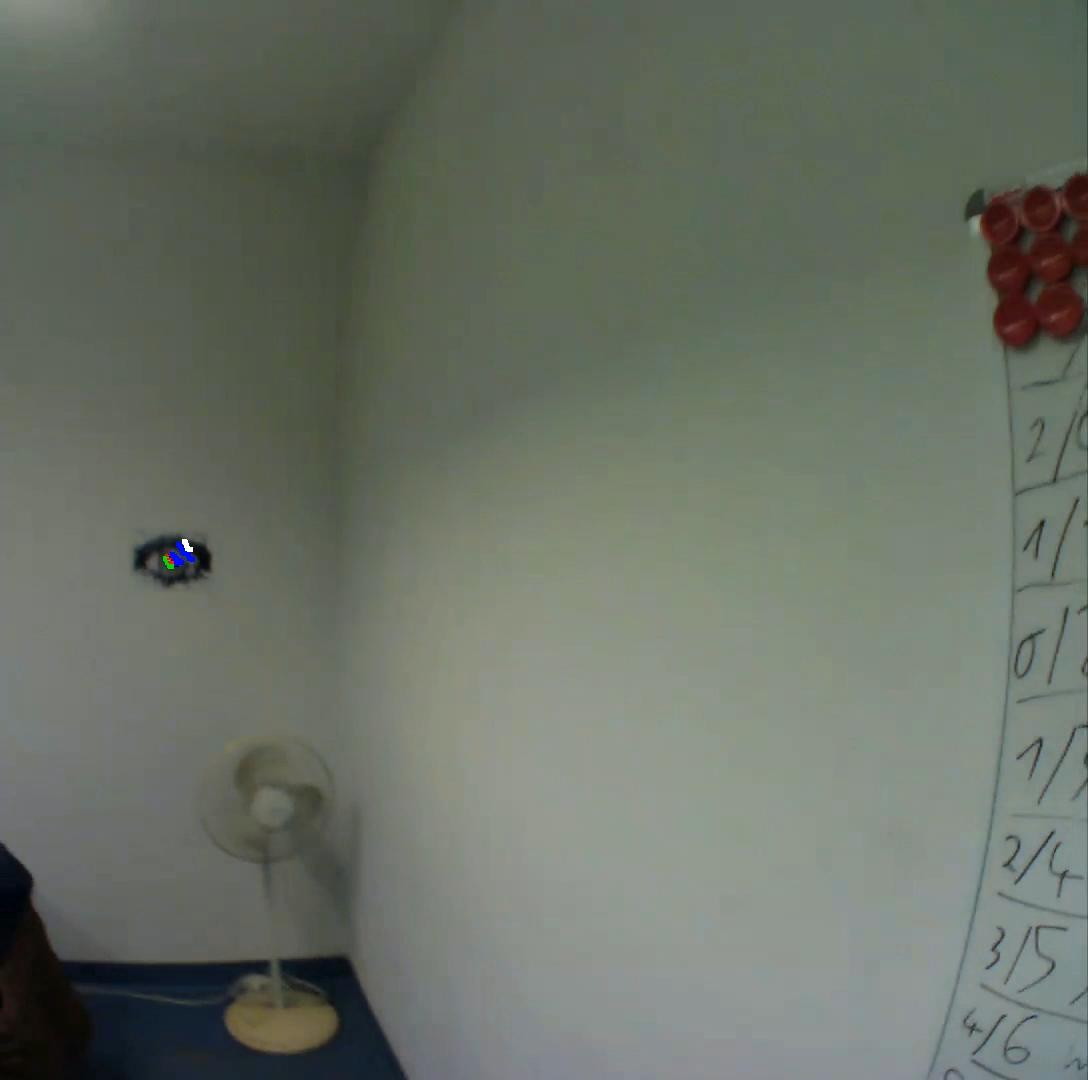}
		\caption{Exemplary images of the 2D gaze estimation for the left (top first 2) and right (top second 2) eye, and exemplary images of the 3D gaze estimation for both eyes (last 4). The iris center, pupil center, iris vector, and pupil vector with neural networks and levenberg marquart polynomial fitting are drawn in different colors. In addition, each scene frame as $\approx 6$ gaze points based on the frame rate of the eye camera.}
		\label{fig:2d3dgaze}
	\end{figure*}
	
	For the determination of the calibration function, we use on the one hand the Levenberg Marquart optimization and on the other hand neural networks with two hidden layers (50 and 20 neurons). This is due to the fact that Pistol should support different eye trackers in the future, and both methods are able to learn more complex functions than simple polynomials. An example of this can already be found in the depth estimation, where the parameters are in the exponent of the function and thus cannot be determined via a direct computation method. 
	
	For the 2D eye tracking we fit a neural network and a polynomial with the Levenberg Marquart method to the pupil center, iris center, pupil vector, and iris vector. This is done separately and also for each eye independently. This means that people with an ocular incorrect eye position or only one functioning eye can also be measured. In the case of 3D eye tracking, we use the data from both eyes simultaneously. This means that we fit a neural network and a polynomial with the Levenberg Marquart method to both pupil centers, iris centers, pupil vectors, and iris vectors. Overall, the program therefore computes 8 gaze point estimators per eye (Total 16) and additionally 8 for both eyes in combination. All of these estimated gaze points can be used separately and are written to the csv files.
	
	The selection of the calibration data is done in three steps. First, each scene image with a marker detection is assigned only one eye image for each eye. This assignment is done by the minimum distance of the time stamps.  In the second step, we check that only scene images are used for which it is true that there are 5 valid marker detections in the preceding and following scene images. In the third step, we use the Levenberg Marquart method together with a polynomial. Based on the error of the gaze positions to the marker position, the best 90\% are selected and the remaining 10\% are discarded.
	
	The training parameters for the neural networks are initial learning rate of $0.1$, optimizer SGD, momentum $0.9$, and weight decay $0.0005$. Each network is trained 4 times for 2000 epochs, with batch size equal to all input data. After 2000 epochs, the learning rate is reduced by a factor of $10^{-1}$. For the Leveberg Marquart method, we use the delta stop strategy with a factor of $1^{-10}$ and a search radius of $10.0$. For the neural networks and the Leveberg Marquart method, all data are normalized. This means that the pupil and iris centers are divided by the resolution in x and y direction, as well as the eyeball centers for the pupil and iris vectors. The vectors themselves are already unit vectors.
	
	\subsection{Depth estimation}
	\label{subsec:depth}
	\begin{figure*}
		\centering
		\includegraphics[width=0.95\textwidth]{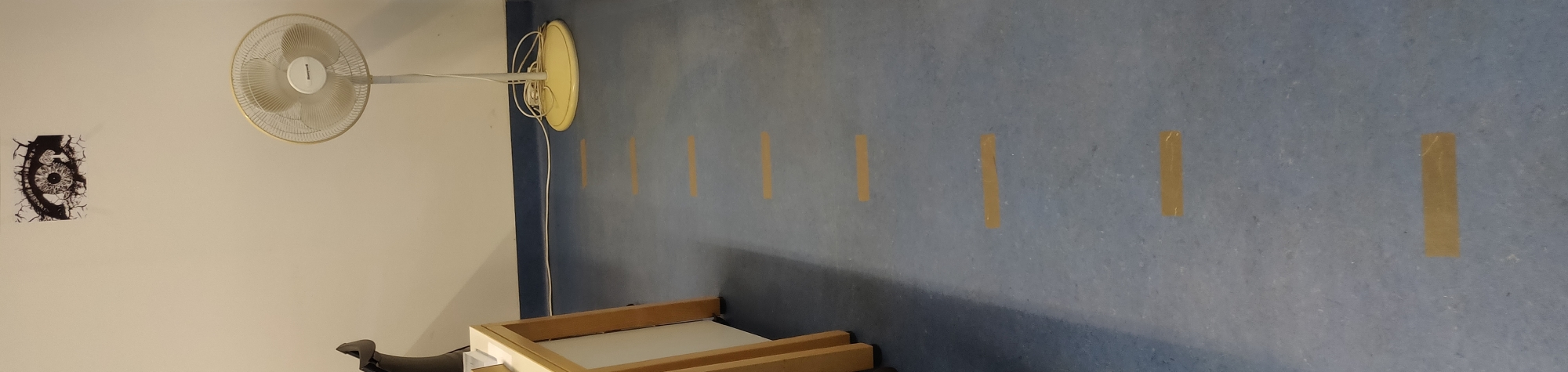}
		\caption{Image of the marker area to depth estimation recording. The brown markers on the floor have 50 cm distance to each other and the subject had to stand in front of the line for one measurement sample.}
		\label{fig:measure}
	\end{figure*}
	For the determination of the depth, we have considered several methods. One is the fitting of polynomials and complex functions to the vectors of the pupil and the iris, as well as to the angle between both vectors. Also, we tried to determine the depth geometrically. In all cases there were significant errors, because the vectors are not linear to each other due to the perspective distortion of the camera, the non-linear deep distribution of the eye image and also due to the head rotation. In the case of the head rotation, it comes to the eye rotation around the z axis as well as it also depends on the viewing angle to different eye positions comes, which are not linear to a straight view. Since neither neural networks nor complex functions worked for us, we decided to use the K nearest neighbor method (KNN) with $k=2$. This method gave the best results in our experiments and a good accuracy (except for a few centimeters).
	
	In order to determine the marker depth and thus obtain our calibration data for the KNN, we performed trial measurements with the marker and two people. The setup can be seen in Figure~\ref{fig:measure}. The brown strokes on the floor are plotted at 50 cm intervals, and both subjects took measurements at each stroke. We then used the Matlab curve fitting toolbox to determine the best function and parameters for them.
	
	\begin{equation}
		\operatorname{d}(\operatorname{A}) = a * \operatorname{A}^{b} + c
		\label{eq:areatodepth}
	\end{equation}
	
	The best function can be seen in the Equation~\ref{eq:areatodepth}, where $A$ is the area of the marker and $\operatorname{d}(A)$ is the depth in centimeters.
	The first thing to notice is that the parameter $b$ is in the exponent, which makes determining the parameter not easy to solve directly. The parameters we determined for the function are $a=13550.0f$, $b=-0.4656$, and $c=-18.02$. We use these parameters and function to determine the depth of the marker from the marker area using the fine detector.
	
	\section{Evaluation}
	In this section we evaluate the different processing steps of the presented tool. If it was possible we compared it to other state of the art approaches. For the gaze estimation we compared to the Pupil Player but we want to mention again, that our tool has to be seen as an additional software to use with the Pupil Invisible and in the future with other eye trackers too. It is not a competing product, nor do we wish to denigrate any software here.
	
	For the evaluation, we recorded five subjects who looked at the calibration marker at the beginning of the recording. Subsequently, the subjects moved freely inside and outside the university and at the end the calibration marker was viewed again. The observation of the calibration marker was at the beginning that the subjects stood close to the marker and then slowly moved away from the marker with head rotations. For the evaluation data, subjects were roughly five meters away initially and then moved toward the marker with head rotations. By doing this, we lift depth information in the calibration as well as in the evaluation data (See section~\ref{subsec:depth}). The head rotation caused the marker to move in the scene image as well as eye rotations, making gaze determination much more difficult. 
	
	Each recording was approximately five minutes long, and each subject took three recordings. One shot could not be used because the subject had the marker in the scene image for a while without looking at it, resulting in a false calibration. Of course, Pistol filters out a certain amount of false data via the outlier selection, but this must not exceed 10\% of the images with markers in the calibration phase. 
	
	Our system for the evaluation consists of an NVIDIA GTX 1050ti, an AMD Ryzen 9 3950X with 16 cores (where only one core was used in the evaluation), and 64 GB DDR4 RAM. The operating system is Windows 10 and the CUDA version used is 11.2, although Pistol also works with CUDA versions 10.X, 11.1, and 11.3.

	\subsection{Pupil, Iris, and Eyelid detection}
	\begin{table}[htb]
		\caption{We evaluated the segmentation quality with the mean intersection over union as well as the average euclidean pupil center error in comparison to other approaches from the literature on our annotated dataset. In addition, we evaluated the iris landmark RMSE also with the euclidean distance. Pistol with 3S stands for the evaluation on three subjects without the two which are part of the annotated data (But from older recordings).}
		\label{tbl:evalPIL}
		\centering
		\begin{tabular}{lccccc}
			\toprule
			Method & Pupil center RMSE (px) & Pupil mIoU & Iris LM RMSE (px) & Iris mIoU & Eyelid mIoU \\
			\midrule
			Pistol & 0.93 & 0.84 & 1.12 & 0.89 & 0.91 \\
			Pistol (3S) & 0.93 & 0.85 & 1.09 & 0.89 & 0.93 \\
			\cite{WTTE032016} & 8.77 & 0.52 & - & - &  - \\
			\cite{WTCKWE092015} & 10.26 & 0.38 & - & - &  - \\
			\cite{WTDTWE092016} & - & - & - & - & 0.53 \\
			\cite{WTE032017} & - & - & - & - & 0.67 \\
			\bottomrule
		\end{tabular}
	\end{table}
	The results of the pupil, iris, and eyelid extraction networks can be seen in Table~\ref{tbl:evalPIL}. As can be seen, the results of Pistol are very good compared to other methods, but the other approaches do not require training data. In the annotated data for feature extraction, there are also two subjects who performed the experiment. The images are from videos that are about one year old. Therefore, we have evaluated the feature extraction in Pistol (3S) additionally only on the subjects that are not in the annotated data. As can be seen, the results are almost identical to the evaluation on all subjects.
	
	Of course, one could use much larger networks with even better results, but Pistol should be usable by everyone, so the resource consumption is also a very important property of the models. In the runtime section, you can see that our approach on an old GTX 1050ti with 4 GB Ram needs only 17.21 milliseconds per frame (Table~\ref{tbl:evalRuntime}) and also uses only a fraction of the GPU RAM. This makes it possible to use Pistol on older hardware.


	\subsection{Eye movement detection}
	\begin{table}[htb]
		\caption{Evaluation of the eye movement detection algorithm in pistol and compared to other state of the art approaches on our small annotated data set.}
		\label{tbl:evalEM}
		\centering
		\begin{tabular}{lcccccccc}
			\toprule
			&\multicolumn{2}{c}{Saccade} & \multicolumn{2}{c}{Fixation} & \multicolumn{2}{c}{Smooth Pursuit} & \multicolumn{2}{c}{Blink}\\
			Method & Accuracy & mIoU & Accuracy & mIoU & Accuracy & mIoU & Accuracy & mIoU \\
			\midrule
			Pistol & 87\% & 0.77 & 96\% & 0.88 & 94\% &  0.87  & 83\% &  0.76  \\
			\cite{ICMIW2019FuhlW2} & 90\% & 0.79 & 94\% & 0.86 & 91\% & 0.82 & 87\% &  0.79  \\
			\cite{ICMIW2019FuhlW1} & 81\% & 0.71 & 90\% & 0.83 & 100\% & 0.84 & 68\% &  0.47  \\
			\cite{FCDGR2020FUHL} & 91\% & 0.84 & 96\% & 0.89 & 97\% & 0.90 & 88\% &  0.81  \\
			\bottomrule
		\end{tabular}
	\end{table}
	Table~\ref{tbl:evalEM} shows the accuracy of our eye movement classification compared to other methods from the literature. The best result is given by the method from \cite{FCDGR2020FUHL} which will soon be integrated into Pistol. However, we want to annotate more data first in order to determine an optimal model via a model search. Regardless, the current approach is better than HOV~\cite{ICMIW2019FuhlW1} and comes close to threshold learning~\cite{ICMIW2019FuhlW2} as used in the classical algorithms. Overall, we are thus very satisfied with the results of Pistol, whereas the bigger challenges for eye movement classification will come mainly from different eye trackers.
	
	\subsection{2D \& 3D Gaze estimation}
	\begin{table}[htb]
		\caption{The average gaze estimation accuracy in pixels for all methods of Pistol. LM stands for the Levenberg-Marquardt fitting, NN for the neural network fitting, PC for pupil center, IC for iris center, PV for the pupil vector, and IV for the iris vector. Since the Pupil Player outputs only up to two gaze points per scene image, we evaluated our approaches with all assigned gaze points to the scene image and only with the eye image with the lowest timestamp difference. The result of 2D left+right is the average of both gaze estimations. \textbf{Note: The Pupil Player uses only the one time calibration for one depth and therefore, the evaluation is not entirely fair.}}
		\label{tbl:evalGaze}
		\centering
		\begin{tabular}{clcccccccc}
			\toprule
			Data & Method & \multicolumn{8}{c}{Average Accuracy (px)} \\
			& & LMPC & LMIC & LMPV & LMIV & NNPC & NNIC & NNPV & NNIV \\
			\midrule
			\multirow{4}{*}{$\frac{\approx 6}{Image}$}& Pistol 2D left & 29.08 & 27.70 & 28.65 & 27.36 & 36.29 & 35.28 & 34.54 & 35.25 \\
			& Pistol 2D right & 32.26 & 33.24 & 34.26 & 35.13 & 47.49 & 44.85 &  37.70 & 38.67\\
			& Pistol 2D left+right & 21.78 & 23.57 & 22.94 & 24.18 & 27.05 & 27.36 & 24.54 & 28.27\\
			& Pistol 3D & 20.26 & 21.22 & 19.61 & 20.48 & 28.78 & 30.29 & 26.83 & 27.90 \\ \hline
			\multirow{4}{*}{$\frac{1}{Image}$}& Pistol 2D left & 24.85 & 23.64 & 24.45 & 23.25 & 32.64 & 31.55 & 30.59 & 31.36 \\
			& Pistol 2D right & 28.07 & 29.07 & 29.96 & 31.00 & 43.97 & 40.84 & 33.84 & 34.62\\
			& Pistol 2D left+right & 18.51 & 20.23 & 19.66 & 20.89 & 24.21 & 24.53 & 21.61 & 25.21\\
			& Pistol 3D & 18.19 & 19.28 & 17.48 & 18.51 & 27.52 & 29.11 & 25.14 & 26.30\\ \hline
			$\frac{\approx 2}{Image}$& PupilLabs & \multicolumn{8}{c}{66.39} \\
			\bottomrule
		\end{tabular}
	\end{table}
	As already described in the introduction in the section Evaluation, we use different depths for calibration and evaluation as well as there are head and eye rotations. Table~\ref{tbl:evalGaze} shows the accuracy of our gaze determination with the different methods. As can be seen, the neural networks are significantly worse than the Levenberg-Marquardt fitting, which we suspect is due to the formulation of the individual neurons. These are single linear functions which are combined to approximate the seen area well and interpolate well between training data. However, in the case of extrapolation, neural networks are very poor. This will be the task of future research, and we hope to provide better networks in the future. Compared to the Pupil Player, we are significantly better on average, but this is also due to the fact that the Pupil Invisible Eye Tracker is only calibrated to one depth level and the software cannot use our markers.
	
	\begin{table}[htb]
		\caption{Since the Pupil Player outputs only one or two gaze points per scene image, we evaluated the total valid gaze points of all possible gaze points (Up to seven per scene image) as well as scene images with at least one gaze point both as percentage. For the evaluation, we used all recordings from all participants.}
		\label{tbl:evalStat}
		\centering
		\begin{tabular}{lcc}
			\toprule
			Source & Valid gaze points & Scene images with at least one valid gaze point\\
			\midrule
			Left & 90.63\% & 91.77\%  \\
			Right & 93.71\% &  94.80\% \\
			Left + Right & 93.78\% & 96.89\%  \\
			3D & 88.14\% &  89.66\% \\
			Pupil Player & 23.07\% & 99.32\%  \\
			\bottomrule
		\end{tabular}
	\end{table}
	Table~\ref{tbl:evalStat} shows the frequency of detected viewpoints relative to all seen images from all shots. As can be seen, it is possible to determine significantly more gaze points with the Pupil Invisible than is provided for in the Pupil Player software. This is due to the fact that the eye cameras record 200 frames per second and the scene camera records 30 frames per second. Looking at the results, it is noticeable that the 3D viewpoints can be calculated the least. This is due to the fact that both eyes must be valid here. However, due to the camera placement, the pupil is no longer visible in one of the eyes when looking from the side. 
	
	If we calculate the relative viewpoints per scene image, i.e. that at least one viewpoint per scene image is sufficient, we see that here the Pupil Player determines significantly more viewpoints than Pistol. For future versions of Pistol, this will be further improved, and we will also integrate an appearance based gaze estimation approach. 
	
	\subsection{Depth estimation}
	\begin{figure*}
		\centering
		\includegraphics[width=0.55\textwidth]{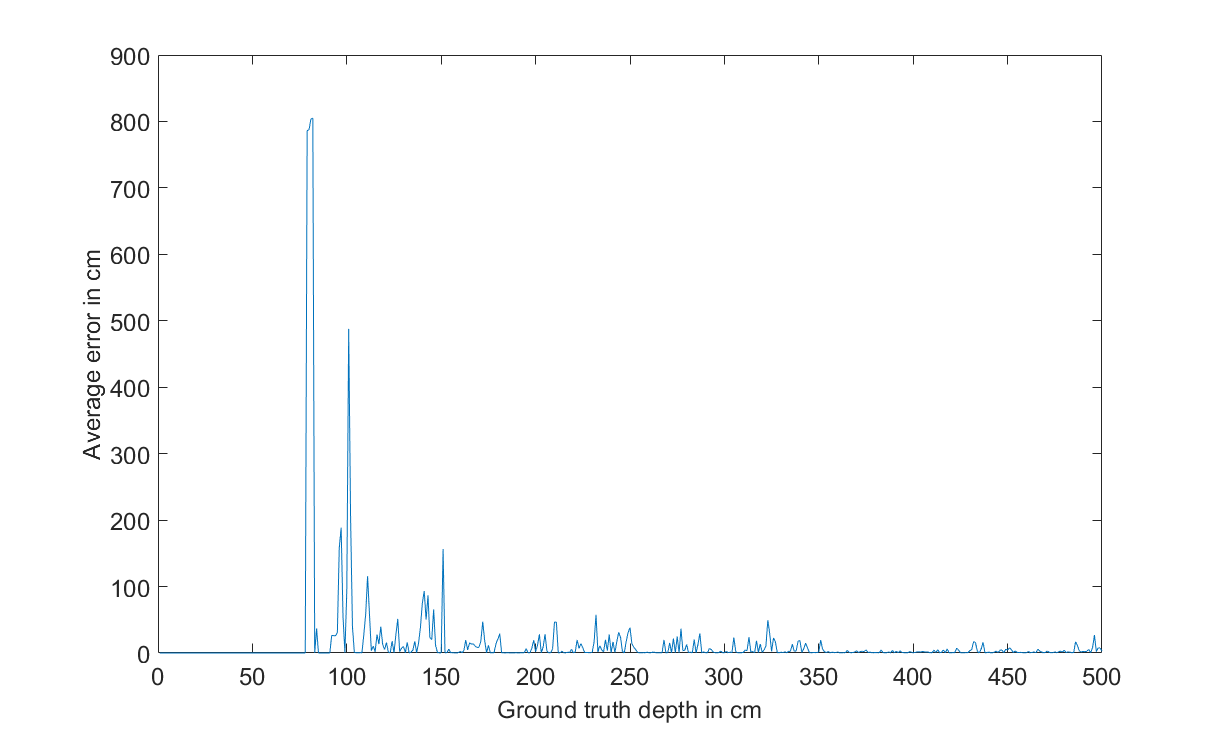}
		\includegraphics[width=0.44\textwidth]{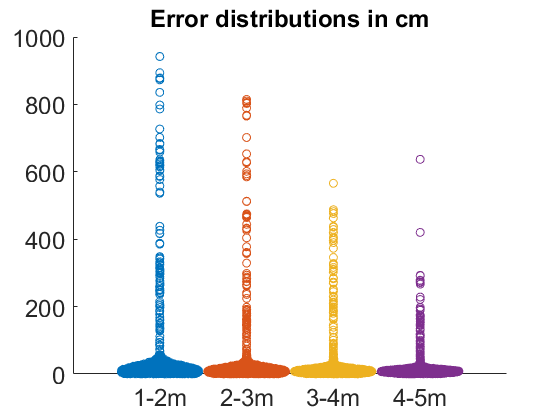}
		\caption{The average error per distance is shown on the right in cm. The left plot is the distribution of errors for the segments in 1 meter blocks. Both plots are computed on all recordings.}
		\label{fig:deptheval}
	\end{figure*}
	Figure~\ref{fig:deptheval} shows the accuracy of our depth estimation. On the left side, the mean values for all depths up to 5 meters are shown. As you can see, especially the near areas are highly error-prone, which is due to the fact that both eyes are very close to the nose, which makes the perspective of the camera placement difficult to detect and distorts the vectors. On the right side, you can see the frequencies in one meter blocks. Here it is clear that there are some outliers, but the most common estimates have an error below 50 cm. This clearly shows that the depth estimation especially for a lateral camera placement needs further research, and we hope to improve it in future versions.

	\subsection{Marker detection}
	\begin{table}[htb]
		\caption{Results of the marker detection. The accuracy and the mean intersection over union is evaluated on our annotated data set. Accuracy is measured as average euclidean distance of the landmarks in pixel. The false detections are evaluated over all recorded videos and divided by the sum of all frames.}
		\label{tbl:evalMarker}
		\centering
		\begin{tabular}{lccc}
			\toprule
			Method & Average landmark Accuracy (px) & mIoU & False detections \\
			\midrule
			Coarse & 6.70 & 0.61 & 2.34\%\\
			Fine & 0.82 & 0.95 & 0.001\% \\
			\bottomrule
		\end{tabular}
	\end{table}
	Table~\ref{tbl:evalMarker} shows the results of our marker detection in the Euclidean distance to the marker landmarks, as mean intersection over union, which is important for the determination of the depth information, and the false detections over all videos. As can be seen, the coarse detection of the markers is relatively error-prone with a false detection rate of 2.34\% as well as very inaccurate. However, this is compensated by the fine detection, and it saves a lot of time to have to check only a few positions with the fine detection.

	\subsection{Runtime}
	\begin{table}[htb]
		\caption{Runtime of each step of Pistol in milliseconds per image or one data instance, without the generation of the debug information and debug video, as average over 1000 data instances. The used GPU is a GTX 1050Ti and the CPU AMD Ryzen 9 3950X where only one core is used for the evaluation.}
		\label{tbl:evalRuntime}
		\centering
		\begin{tabular}{lc}
			\toprule
			Step & Runtime in ms\\
			\midrule
			Feature detection & 17.21 \\
			Eye ball and vector & 0.22 \\
			Eye opening & 0.18 \\
			Eye movements & 0.49 \\
			Synchronisation & 0.01 \\
			Marker detection & 57.35 \\
			Gaze estimation LM & 0.14 \\
			Gaze estimation NN & 0.27 \\
			Training LM Gaze & 2958 \\
			Training NN Gaze & 6107 \\
			\bottomrule
		\end{tabular}
	\end{table}
	Table~\ref{tbl:evalRuntime} shows the runtime of the individual steps of Pistol. The times are in milliseconds and include the use of the GPU for marker detection and feature extraction. In the first update, the use of Pistol will also be possible without GPU, and thus we can then also provide a version for MacOSX. The training for the gaze determination takes the most time, however, it is not only a data instance, but the complete training with all data. In the future, the marker detection will be accelerated as well, since it currently still takes too much time. However, it was important to us that the marker detection is very accurate and robust.

	\section{Limitations}
	The limitations of our tool are currently that we only support the Pupil Invisible, that we can only offer the software for Windows and Linux, that a CUDA capable GPU is required, and that we so far only support one marker. In the future, we plan to add more eye trackers, but the prerequisite is that we get access to the eyes and the scene video. We would also like to provide an application for MacOSX, but unfortunately we don't have the hardware to test the whole system. For the usability of our software without a CUDA capable GPU the implementation is already done, but smaller models need to be trained and pruned for real-time processing on CPUs like the TinyCNNs from \cite{VECETRA2020}. Once the smaller models are finalized, we will make our software available without the need for a GPU. Further, updates of our software will allow different markers to be used as well as a gesture recognition model will be integrated which can then of course also be used for calibration.
	
	\section{Conclusion}
	In this paper we have described in detail our software that allows to extract a variety of features from the eye as well as to perform 2D and 3D gaze estimation. The software still has certain limitations such as the need for a CUDA capable GPU and the restriction to one marker, but these limitations will soon be removed, and other eye trackers will be added. We hope that the variety of extracted features will support research in many areas and that the software will be useful for many researchers. Future features of the software will be gesture recognition, scene analysis as well as scan path classification, which will hopefully make the software even more useful for research.

	\bibliographystyle{plain}
	\bibliography{template}

\begin{thebibliography}{10}

\bibitem{arslan2021eye}
Omer Arslan, Oguz Atik, and Serkan Kahraman.
\newblock Eye tracking in usability of electronic chart display and information
  system.
\newblock {\em The Journal of Navigation}, 74(3):594--604, 2021.

\bibitem{ASL}
ASL.
\newblock Asl.
\newblock \url{https://est-kl.com/}, 2021.
\newblock [Online; accessed 04-November-2021].

\bibitem{benjamins2018gazecode}
Jeroen~S Benjamins, Roy~S Hessels, and Ignace~TC Hooge.
\newblock Gazecode: Open-source software for manual mapping of mobile
  eye-tracking data.
\newblock In {\em Proceedings of the 2018 ACM symposium on eye tracking
  research \& applications}, pages 1--4, 2018.

\bibitem{bradski2000opencv}
Gary Bradski.
\newblock The opencv library.
\newblock {\em Dr. Dobb's Journal: Software Tools for the Professional
  Programmer}, 25(11):120--123, 2000.

\bibitem{davis2020eye}
Rebecca Davis and Alla Sikorskii.
\newblock Eye tracking analysis of visual cues during wayfinding in early stage
  alzheimer’s disease.
\newblock {\em Dementia and geriatric cognitive disorders}, 49(1):91--97, 2020.

\bibitem{duchowski2002breadth}
Andrew~T Duchowski.
\newblock A breadth-first survey of eye-tracking applications.
\newblock {\em Behavior Research Methods, Instruments, \& Computers},
  34(4):455--470, 2002.

\bibitem{economides2007ocular}
John~R Economides, Daniel~L Adams, Cristina~M Jocson, and Jonathan~C Horton.
\newblock Ocular motor behavior in macaques with surgical exotropia.
\newblock {\em Journal of neurophysiology}, 98(6):3411--3422, 2007.

\bibitem{Ergoneers}
Ergoneers.
\newblock Ergoneers.
\newblock \url{https://www.ergoneers.com}, 2021.
\newblock [Online; accessed 04-November-2021].

\bibitem{eyecomtec}
eyecomtec.
\newblock eyecomtec.
\newblock \url{https://eyecomtec.com}, 2021.
\newblock [Online; accessed 04-November-2021].

\bibitem{EyeTech}
EyeTech.
\newblock Eyetech.
\newblock \url{http://www.eyetec.com/}, 2021.
\newblock [Online; accessed 04-November-2021].

\bibitem{ICMIW2019FuhlW1}
W.~Fuhl, N.~Castner, and E.~Kasneci.
\newblock Histogram of oriented velocities for eye movement detection.
\newblock In {\em International Conference on Multimodal Interaction Workshops,
  ICMIW}, 2018.

\bibitem{ICMIW2019FuhlW2}
W.~Fuhl, N.~Castner, and E.~Kasneci.
\newblock Rule based learning for eye movement type detection.
\newblock In {\em International Conference on Multimodal Interaction Workshops,
  ICMIW}, 2018.

\bibitem{NNETRA2020}
W.~Fuhl, H.~Gao, and E.~Kasneci.
\newblock Neural networks for optical vector and eye ball parameter estimation.
\newblock In {\em ACM Symposium on Eye Tracking Research \& Applications, ETRA
  2020}. ACM, 01 2020.

\bibitem{VECETRA2020}
W.~Fuhl, H.~Gao, and E.~Kasneci.
\newblock Tiny convolution, decision tree, and binary neuronal networks for
  robust and real time pupil outline estimation.
\newblock In {\em ACM Symposium on Eye Tracking Research \& Applications, ETRA
  2020}. ACM, 01 2020.

\bibitem{ICMV2019FuhlW}
W.~Fuhl and E.~Kasneci.
\newblock Learning to validate the quality of detected landmarks.
\newblock In {\em International Conference on Machine Vision, ICMV}, 11 2019.

\bibitem{WTCKWE092015}
W.~Fuhl, T.~C. Kübler, K.~Sippel, W.~Rosenstiel, and E.~Kasneci.
\newblock Excuse: Robust pupil detection in real-world scenarios.
\newblock In {\em 16th International Conference on Computer Analysis of Images
  and Patterns (CAIP 2015)}, 09 2015.

\bibitem{WTDTWE092016}
W.~Fuhl, T.~Santini, D.~Geisler, T.~C. Kübler, W.~Rosenstiel, and E.~Kasneci.
\newblock Eyes wide open? eyelid location and eye aperture estimation for
  pervasive eye tracking in real-world scenarios.
\newblock In {\em ACM International Joint Conference on Pervasive and
  Ubiquitous Computing: Adjunct publication -- PETMEI 2016}, 09 2016.

\bibitem{WTE032017}
W.~Fuhl, T.~Santini, and E.~Kasneci.
\newblock Fast and robust eyelid outline and aperture detection in real-world
  scenarios.
\newblock In {\em IEEE Winter Conference on Applications of Computer Vision
  (WACV 2017)}, 03 2017.

\bibitem{WTTE032016}
W.~Fuhl, T.~Santini, T.~C. Kübler, and E.~Kasneci.
\newblock Else: Ellipse selection for robust pupil detection in real-world
  environments.
\newblock In {\em Proceedings of the Ninth Biennial ACM Symposium on Eye
  Tracking Research \& Applications (ETRA)}, pages 123--130, 03 2016.

\bibitem{NIPS2021MAXPROP}
Wolfgang Fuhl.
\newblock Maximum and leaky maximum propagation.
\newblock {\em arXiv preprint arXiv:2105.10277}, 2021.

\bibitem{2021TNandFDT}
Wolfgang Fuhl.
\newblock Tensor normalization and full distribution training.
\newblock {\em arXiv preprint arXiv:2109.02345}, 2021.

\bibitem{FCDGR2020FUHL}
Wolfgang Fuhl, Yao Rong, and Kasneci Enkelejda.
\newblock Fully convolutional neural networks for raw eye tracking data
  segmentation, generation, and reconstruction.
\newblock In {\em Proceedings of the International Conference on Pattern
  Recognition}, pages 0--0, 2020.

\bibitem{gardony2020eye}
Aaron~L Gardony, Robert~W Lindeman, and Tad~T Bruny{\'e}.
\newblock Eye-tracking for human-centered mixed reality: promises and
  challenges.
\newblock In {\em Optical Architectures for Displays and Sensing in Augmented,
  Virtual, and Mixed Reality (AR, VR, MR)}, volume 11310, page 113100T.
  International Society for Optics and Photonics, 2020.

\bibitem{hale2019eyestream}
Matthew~L Hale.
\newblock Eyestream: An open websocket-based middleware for serializing and
  streaming eye tracker event data from gazepoint gp3 hd research hardware.
\newblock {\em Journal of Open Source Software}, 4(43):1620, 2019.

\bibitem{hasselbring2020open}
Wilhelm Hasselbring, Leslie Carr, Simon Hettrick, Heather Packer, and Thanassis
  Tiropanis.
\newblock Open source research software.
\newblock {\em Computer}, 53(8):84--88, 2020.

\bibitem{he2016deep}
Kaiming He, Xiangyu Zhang, Shaoqing Ren, and Jian Sun.
\newblock Deep residual learning for image recognition.
\newblock In {\em Proceedings of the IEEE conference on computer vision and
  pattern recognition}, pages 770--778, 2016.

\bibitem{iMotions}
iMotions.
\newblock imotions.
\newblock \url{https://imotions.com/}, 2021.
\newblock [Online; accessed 04-November-2021].

\bibitem{jermann2010using}
Patrick Jermann, Marc-Antoine N{\"u}ssli, and Weifeng Li.
\newblock Using dual eye-tracking to unveil coordination and expertise in
  collaborative tetris.
\newblock {\em Proceedings of HCI 2010 24}, pages 36--44, 2010.

\bibitem{jiang2020design}
Zongmin Jiang, Yan Chang, and Xuefen Liu.
\newblock Design of software-defined gateway for industrial interconnection.
\newblock {\em Journal of Industrial Information Integration}, 18:100130, 2020.

\bibitem{jones2018myex}
Pete~Richard Jones.
\newblock Myex: a matlab interface for the tobii eyex eye-tracker.
\newblock {\em Journal of Open Research Software}, 6(1), 2018.

\bibitem{joseph2020potential}
Antony~William Joseph and Ramaswamy Murugesh.
\newblock Potential eye tracking metrics and indicators to measure cognitive
  load in human-computer interaction research.
\newblock {\em Journal of Scientific Research}, 64(1), 2020.

\bibitem{lepekhin2020adoption}
Aleksander Lepekhin, Dorothea Capo, Anastasia Levina, Alexandra Borremans, and
  Zarema Khasheva.
\newblock Adoption of industrie 4.0 technologies in the manufacturing companies
  in russia.
\newblock In {\em Proceedings of the International Scientific
  Conference-Digital Transformation on Manufacturing, Infrastructure and
  Service}, pages 1--6, 2020.

\bibitem{lev2020eye}
Astar Lev, Yoram Braw, Tomer Elbaum, Michael Wagner, and Yuri Rassovsky.
\newblock Eye tracking during a continuous performance test: Utility for
  assessing adhd patients.
\newblock {\em Journal of Attention Disorders}, page 1087054720972786, 2020.

\bibitem{li2006openeyes}
Dongheng Li, Jason Babcock, and Derrick~J Parkhurst.
\newblock openeyes: a low-cost head-mounted eye-tracking solution.
\newblock In {\em Proceedings of the 2006 symposium on Eye tracking research \&
  applications}, pages 95--100, 2006.

\bibitem{liu2019gaze}
Congcong Liu, Yuying Chen, Lei Tai, Haoyang Ye, Ming Liu, and Bertram~E Shi.
\newblock A gaze model improves autonomous driving.
\newblock In {\em Proceedings of the 11th ACM symposium on eye tracking
  research \& applications}, pages 1--5, 2019.

\bibitem{loshchilov2018fixing}
Ilya Loshchilov and Frank Hutter.
\newblock Fixing weight decay regularization in adam.
\newblock 2018.

\bibitem{mao2021survey}
Runze Mao, Guoyuan Li, Hans~Petter Hildre, and Houxiang Zhang.
\newblock A survey of eye tracking in automobile and aviation studies:
  Implications for eye-tracking studies in marine operations.
\newblock {\em IEEE Transactions on Human-Machine Systems}, 51(2):87--98, 2021.

\bibitem{meng2020eye}
Xiaoxu Meng, Ruofei Du, and Amitabh Varshney.
\newblock Eye-dominance-guided foveated rendering.
\newblock {\em IEEE transactions on visualization and computer graphics},
  26(5):1972--1980, 2020.

\bibitem{nesaratnam2017stepping}
Nisha Nesaratnam, Peter Thomas, and Anthony Vivian.
\newblock Stepping into the virtual unknown: feasibility study of a virtual
  reality-based test of ocular misalignment.
\newblock {\em Eye}, 31(10):1503--1506, 2017.

\bibitem{niehorster2020glassesviewer}
Diederick~C Niehorster, Roy~S Hessels, and Jeroen~S Benjamins.
\newblock Glassesviewer: Open-source software for viewing and analyzing data
  from the tobii pro glasses 2 eye tracker.
\newblock {\em Behavior Research Methods}, 52(3):1244--1253, 2020.

\bibitem{Oculus}
Oculus.
\newblock Oculus.
\newblock \url{https://www.oculus.com}, 2021.
\newblock [Online; accessed 04-November-2021].

\bibitem{panchuk2015eye}
Derek Panchuk, Samuel Vine, and Joan~N Vickers.
\newblock Eye tracking methods in sport expertise.
\newblock In {\em Routledge handbook of sport expertise}, pages 176--187.
  Routledge, 2015.

\bibitem{park2021technical}
Junghoon Park and Kangbin Yim.
\newblock Technical survey on the real time eye-tracking pointing device as a
  smart medical equipment.
\newblock {\em Smart Media Journal}, 10(1):9--15, 2021.

\bibitem{pavisic2021eye}
Ivanna~M Pavisic, Yoni Pertzov, Jennifer~M Nicholas, Antoinette O’Connor,
  Kirsty Lu, Keir~XX Yong, Masud Husain, Nick~C Fox, and Sebastian~J Crutch.
\newblock Eye-tracking indices of impaired encoding of visual short-term memory
  in familial alzheimer’s disease.
\newblock {\em Scientific reports}, 11(1):1--14, 2021.

\bibitem{phan2011development}
Truong~Vinh Phan.
\newblock {\em Development of a custom application for the Tobii Eye Tracker}.
\newblock PhD thesis, Hochschule f{\"u}r angewandte Wissenschaften Hamburg,
  2011.

\bibitem{PupilLabs}
PupilLabs.
\newblock Pupillabs.
\newblock \url{https://pupil-labs.com/}, 2021.
\newblock [Online; accessed 04-November-2021].

\bibitem{rakhmatulin2020review}
Ildar Rakhmatulin.
\newblock A review of the low-cost eye-tracking systems for 2010-2020.
\newblock {\em arXiv preprint arXiv:2010.05480}, 2020.

\bibitem{rumelhart1986learning}
David~E Rumelhart, Geoffrey~E Hinton, and Ronald~J Williams.
\newblock Learning representations by back-propagating errors.
\newblock {\em nature}, 323(6088):533--536, 1986.

\bibitem{santini2017eyerectoo}
Thiago Santini, Wolfgang Fuhl, David Geisler, and Enkelejda Kasneci.
\newblock Eyerectoo: Open-source software for real-time pervasive head-mounted
  eye tracking.
\newblock In {\em VISIGRAPP (6: VISAPP)}, pages 96--101, 2017.

\bibitem{shinohara2017visual}
Yumiko Shinohara, Rebecca Currano, Wendy Ju, and Yukiko Nishizaki.
\newblock Visual attention during simulated autonomous driving in the us and
  japan.
\newblock In {\em Proceedings of the 9th International Conference on Automotive
  User Interfaces and Interactive Vehicular Applications}, pages 144--153,
  2017.

\bibitem{snell2020assessment}
Sarah Snell, Daniel Bontempo, Gregory Celine, and Robert Anthonappa.
\newblock Assessment of medical practitioners' knowledge about paediatric oral
  diagnosis and gaze patterns using eye tracking technology.
\newblock {\em International Journal of Paediatric Dentistry}, 2020.

\bibitem{SRResearch}
SRResearch.
\newblock Srresearch.
\newblock \url{https://www.sr-research.com/}, 2021.
\newblock [Online; accessed 04-November-2021].

\bibitem{taylor2020operator}
Mark~P Taylor, Peter Boxall, John~JJ Chen, Xun Xu, Angela Liew, and Adebayo
  Adeniji.
\newblock Operator 4.0 or maker 1.0? exploring the implications of industrie
  4.0 for innovation, safety and quality of work in small economies and
  enterprises.
\newblock {\em Computers \& industrial engineering}, 139:105486, 2020.

\bibitem{Tobii}
Tobii.
\newblock Tobii.
\newblock \url{https://www.tobii.com/}, 2021.
\newblock [Online; accessed 04-November-2021].

\bibitem{VIVE}
VIVE.
\newblock Vive.
\newblock \url{https://www.vive.com/}, 2021.
\newblock [Online; accessed 04-November-2021].

\bibitem{vosskuhler2008ogama}
Adrian Vo{\ss}k{\"u}hler, Volkhard Nordmeier, Lars Kuchinke, and Arthur~M
  Jacobs.
\newblock Ogama (open gaze and mouse analyzer): open-source software designed
  to analyze eye and mouse movements in slideshow study designs.
\newblock {\em Behavior research methods}, 40(4):1150--1162, 2008.

\bibitem{walton2021beyond}
David~R Walton, Rafael Kuffner~Dos Anjos, Sebastian Friston, David Swapp, Kaan
  Ak{\c{s}}it, Anthony Steed, and Tobias Ritschel.
\newblock Beyond blur: Real-time ventral metamers for foveated rendering.
\newblock {\em ACM Transactions on Graphics (TOG)}, 40(4):1--14, 2021.

\bibitem{zandi2021pupilext}
Babak Zandi, Moritz Lode, Alexander Herzog, Georgios Sakas, and Tran~Quoc
  Khanh.
\newblock Pupilext: Flexible open-source platform for high-resolution
  pupillometry in vision research.
\newblock {\em Frontiers in neuroscience}, 15, 2021.

\end{thebibliography}

\end{document}